%% file: main_redacted.tex
\documentclass{article}

\PassOptionsToPackage{numbers,compress}{natbib}

\usepackage[preprint]{neurips_2026}

\usepackage[utf8]{inputenc} 
\usepackage[T1]{fontenc}    
\usepackage{hyperref}       
\usepackage{url}            
\usepackage{booktabs}       
\usepackage{amsfonts}       
\usepackage{amsmath}        
\usepackage{nicefrac}       
\usepackage{microtype}      
\usepackage[table]{xcolor}  
\usepackage{colortbl}       
\usepackage{graphicx}       
\usepackage{rotating}       
\usepackage{longtable}      
\usepackage{array}          
\usepackage{pdflscape}      
\usepackage{multirow}
\usepackage{caption}        
\definecolor{tablegray}{gray}{0.93}
\definecolor{shadegray}{gray}{0.93}  

\title{Robustifying pathology foundation models via fine-tuning}

\author{%
  Alexandre Filiot\textsuperscript{1}\thanks{Corresponding author: \texttt{alexandre.filiot@wearewaiv.com}} \quad
  Oskar Thaeter\textsuperscript{1,2,3} \quad
  Benoît Schmauch\textsuperscript{1} \quad
  Lionel Guillou\textsuperscript{1} \\[0.5em]
  \textsuperscript{1}Waiv \\
  \textsuperscript{2}Institute of Pathology, Technical University of Munich \\
  \textsuperscript{3}School of Computation, Information and Technology, Technical University of Munich \\[0.5em]
}

\begin{document}

\maketitle

\begin{abstract}
Pathology foundation models (FMs) produce powerful tile-level representations which remain sensitive to scanner and staining variability, undermining
deployment across laboratories. We develop a novel fine-tuning recipe that
improves the robustness of pathology FMs to acquisition factors. Applied to ten different FMs, our fine-tuning strategy consistently improves robustness for every model as well as downstream performance, with no observed trade-off. On average, it raises the PathoROB robustness index by 23\% (from 0.72 to 0.87) and increases the overall cross-benchmark performance by 43\% on Patho-Bench, HEST and THUNDER combined, with individual gains reaching up to 72\% in robustness (Phikon-v2) and 76\% in performance (Midnight-12k). We publicly release the fine-tuned versions of Phikon-v2 (\textit{Phaet}) and Midnight-12k (\textit{Mascaret}) at \href{https://huggingface.co/wearewaiv/models}{huggingface.co/wearewaiv/models}.

\end{abstract}

\begin{center}
  \includegraphics[width=0.62\linewidth]{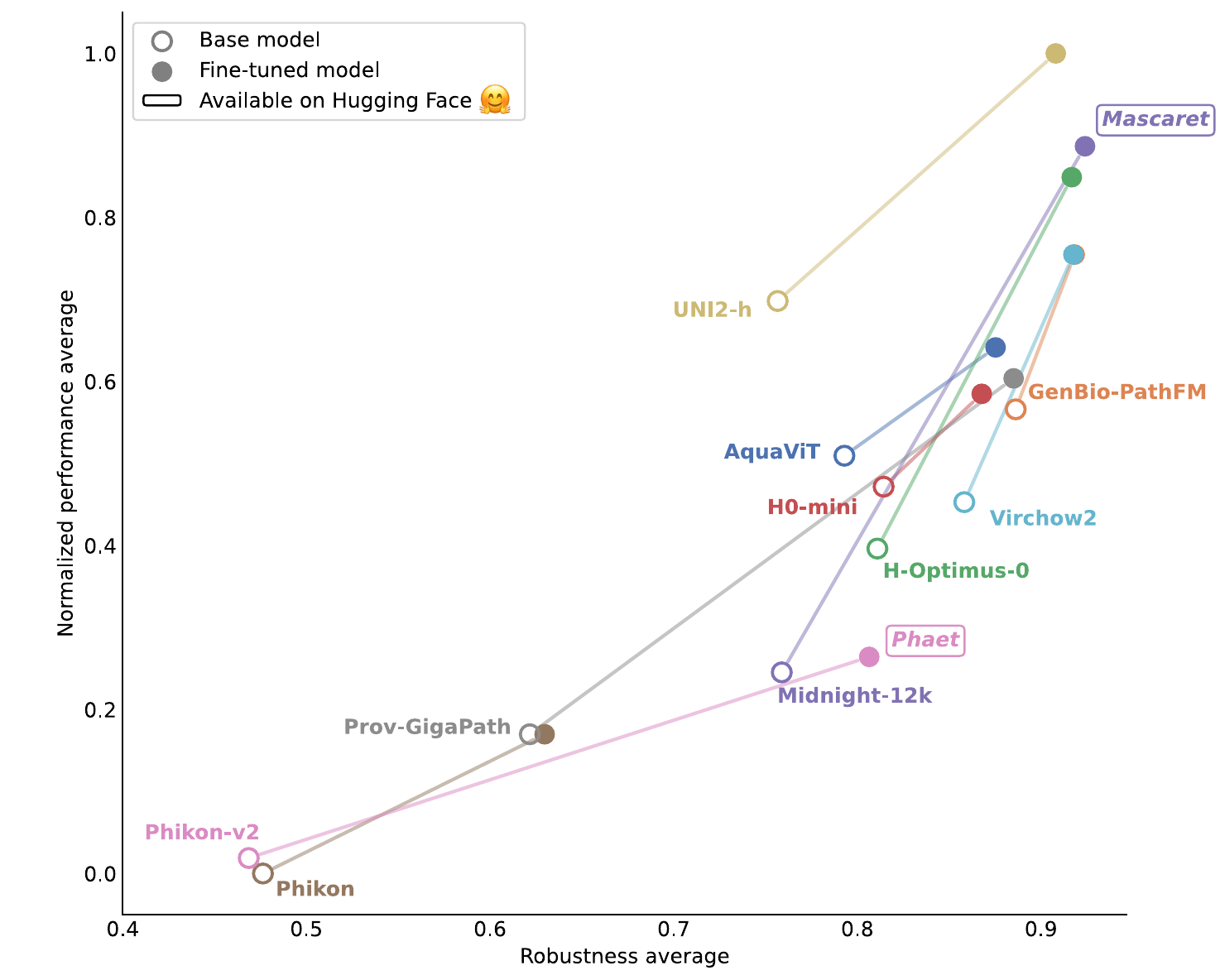}
  \captionof{figure}{\textbf{Fine-tuning improves robustness and performance jointly.} Each
  foundation model is shown before (open circle) and after (filled circle) fine-tuning.
  The $x$-axis is the average PathoROB robustness index over three datasets, where higher values indicate greater robustness; the $y$-axis is the normalized rank sum over the HEST, THUNDER and Patho-Bench benchmarks, rescaled to $[0, 1]$ so that 1 corresponds to the best achievable performance.}
  \label{fig:abstract}
\end{center}

\section{Introduction}
\label{sec:intro}

Foundation models (FMs) trained on large corpora of histopathology images now underpin a wide
range of computational pathology pipelines, involving biomarker prediction~\citep{kather2019msi, saillard2023msintuit}, gene expression prediction~\citep{schmauch2020rnaseq, jaume2024hest}, whole slide image (WSI) and tissue classification~\citep{shao2021transmil, ding2025titan} or survival analysis~\citep{courtiol2019meso, wang2024chief}. Despite their strong performance, these models inherit a long-standing
weakness of the field: their representations are entangled with acquisition factors
such as the scanner model or the staining protocol \cite{dejong2025unrobust}. Two images of the same tissue, scanned
on different devices or stained in different laboratories, can map to markedly different
points in feature space. The resulting domain shift degrades downstream models when they are
deployed in a laboratory whose acquisition pipeline differs from the training distribution
\citep{gustafsson2024evaluating, komen2026robust,scanner2026thiringer}.

We argue that robustness to these confounding factors can be instilled directly into the
encoder, without retraining from scratch and without sacrificing the generic
quality of the representation.

Our contributions are threefold:
\begin{itemize}
  \item We develop a novel fine-tuning recipe
  that makes pathology FMs more robust to acquisition factors, applied
  uniformly to ten different foundation models.
  \item We support our method with a comprehensive empirical study spanning robustness (PathoROB)~\citep{komen2026robust} and downstream performance (HEST gene-expression prediction~\citep{jaume2024hest}, THUNDER tile-level tasks~\citep{marza2025thunder} and Patho-Bench slide-level tasks~\citep{zhang2025pathobench}), showing consistent robustness gains with no performance trade-off (Figure~\ref{fig:abstract}).
  \item We release \textit{Phaet} and \textit{Mascaret}, the respective robust versions of Phikon-v2 and Midnight-12k at \href{https://huggingface.co/wearewaiv/models}{huggingface.co/wearewaiv/models}. Mascaret strikes a state-of-the-art balance between robustness (first on PathoROB) and downstream performance (second on average behind UNI2-h) among the publicly available models.
\end{itemize}

\section{Related work}
\label{sec:related}

Pretraining a feature extractor with self-supervised learning (SSL) is now a cornerstone of modern computational pathology (CPath) frameworks. Early CPath pipelines used to leverage models pretrained on ImageNet~\citep{deng2009imagenet}, and suffered from an out-of-domain gap when transferred to digital
pathology images. The advent of SSL methods showed the benefit of doing an in-domain pretraining for pathology feature extractors~\citep{ciga2021contrastive, dehaene2020selfsup, saillard2021ssl}. While they were
originally leveraging contrastive methods~\citep{chen2020mocov2, zbontar2021barlow} tailored for convolutional networks, more recent models build on the Vision Transformer (ViT) architecture, with pretraining methods such as DINO, iBOT, DINOv2 or DINOv3~\citep{caron2021dino, zhou2022ibot, oquab2024dinov2, simeoni2025dinov3}. Most of the recent foundation models for digital pathology leverage the DINOv2~\citep{ma2026gpfm, saillard2024hoptimus, wolflein2025benchmark, xu2024provgigapath} and DINOv3~\citep{genbiopathfm} frameworks. Over the last two years, progress has largely been driven by scaling the pre-training corpus and model
size, culminating in billion-parameter models trained on millions of WSIs~\citep{zimmermann2024virchow2, alber2026atlas2,hoptimus1,saillard2024hoptimus}. As a consequence of large data diversity scaling, those large models are not only more performant but also more robust to acquisition shift without any explicit invariance mechanism other than the SSL pretext tasks. However, a growing body of work has shown that even recent FMs entangle non-biological acquisition factors (scanner, laboratory, staining) in their
representations, degrading downstream performance across sites~\citep{dejong2025unrobust,
lin2025institution, chai2026impact, gustafsson2024evaluating, howard2021site, dehkharghanian2023biased, tizhoosh2025beyond}, and dedicated benchmarks now quantify this fragility~\citep{komen2026robust,filiot2025distill}. This suggests that perfect robustness is not yet achieved, and the question of how to instill invariance to acquisition factors in a model-agnostic way remains open.

Existing mitigation strategies act away from the encoder. Image-level approaches normalize stain appearance~\citep{macenko2009, reinhard2001, vahadane2016structure} or augment
it~\citep{drexlin2025medi, tellez2019quantifying, shen2022randstainna, nguyen2023contrimix, jahanifar2025domain}; post-hoc approaches correct features using site
statistics~\citep{nguyen2025fmmap}; and downstream-model approaches suppress domain-predictive
features during task training, via domain-adversarial objectives~\citep{ganin2016dann}, contrastive losses over co-registered scanner pairs~\citep{carloni2025scangen,
henriksen2026enabling, ryu2025scorpion}, or information-bottleneck adaptation that disentangles site and demographic artifacts~\citep{huang2025flex}. These downstream methods all keep the backbone frozen. Distillation~\citep{filiot2025distill, ma2026gpfm, grashei2025pathryoshka} has also been proposed to improve efficiency and robustness but requires a robust teacher model and is computationally expensive.

In contrast, we robustify the encoder itself by fine-tuning it, yielding a single task-agnostic and robust encoder.

\section{Experimental setup}
\label{sec:setup}

\subsection{Backbones}
We fine-tune ten pathology FMs spanning a range of architectures and pre-training recipes: AquaViT (internal, ViT-B/14), GenBio-PathFM~\citep{genbiopathfm}, H-Optimus-0~\citep{saillard2024hoptimus}, H0-mini~\citep{filiot2025distill}, Midnight-12k~\citep{karasikov2025midnight}, Phikon~\citep{filiot2023phikon}, Phikon-v2~\citep{filiot2024phikonv2}, Prov-GigaPath~\citep{xu2024provgigapath}, UNI2-h~\citep{chen2024uni}, and Virchow2~\citep{zimmermann2024virchow2}. For detailed information on each model, refer to \href{https://wearewaiv.github.io/histoboard/models}{wearewaiv.github.io/histoboard/models}. The fine-tuned versions of Phikon-v2 and Midnight-12k are referred to as \textit{Phaet} and \textit{Mascaret}, respectively.

\subsection{Robustness evaluation}

\paragraph{PathoROB} We evaluate robustness with the PathoROB benchmark~\citep{komen2026robust}, which introduces the robustness index metric to quantify the robustness of pathology foundation models to non-biological confounding features, specifically
medical center differences arising from variations in staining procedures, scanner hardware, surgical techniques, and laboratory protocols. The Robustness Index (RI) measures the degree to which biological features (e.g., tissue type, cancer type) dominate over confounding non-biological features (e.g.,
medical center signatures) in the neighborhood structure of a foundation model’s feature space. The benchmark comprises three datasets sourced from CAMELYON~\citep{bejnordi2017camelyon16, bandi2019camelyon17}, TCGA~\citep{komura2022tcga} and Tolkach ESCA~\citep{tolkach2023esca}, covering 28 biological classes from 34 medical centers.

\subsection{Performance evaluation}

\paragraph{HEST} The HEST-Benchmark~\citep{jaume2024hest} casts spatial gene-expression prediction as a multivariate regression from tile features, spanning nine cancer types. Following the default protocol and implementation, we fit a ridge regression on PCA-reduced foundation-model features ($d=256$) to predict the measured expression, and report the Pearson correlation coefficient between predicted and measured gene expression. Results are reported for each of the following nine cancer types: invasive ductal carcinoma (breast cancer, IDC),
prostate adenocarcinoma (prostate cancer, PRAD), pancreatic adenocarcinoma (pancreatic cancer, PAAD), skin cutaneous melanoma (skin cancer, SKCM), colonic adenocarcinoma (colon cancer, COAD), rectal adenocarcinoma (rectum cancer, READ), clear cell renal cell carcinoma (kidney cancer, ccRCC), hepatocellular carcinoma (liver cancer, HCC), and axillary lymph nodes in IDC (metastatic, LYMPH-IDC).

\paragraph{THUNDER} THUNDER~\citep{marza2025thunder} is a comprehensive tile-level benchmark designed to rigorously compare foundation models across various downstream tasks in computational pathology. Across 16 different datasets of various sizes, organs and magnifications, THUNDER evaluates $k$-nearest-neighbor classification, linear probing and few-shot (simple-shot) classification, semantic segmentation, predictive calibration, and robustness to adversarial attacks. Each model is evaluated on frozen features following the default protocol and ranked per task; we summarize overall performance by the rank sum across the six tasks following the original leaderboard. For computational efficiency, all THUNDER experiments use mixed precision.

\paragraph{Patho-Bench} Patho-Bench~\citep{zhang2025pathobench} is a slide-level benchmarking framework spanning 95 tasks across seven categories: morphological subtyping, TME characterization, tumor grading, molecular subtyping, mutation prediction, treatment-response assessment, and survival prediction. We evaluate a subset of 63 tasks drawn from multiple datasets under the default ABMIL~\citep{ilse2018abmil} protocol, and report the task-appropriate metric defined by Patho-Bench: AUROC for binary classification, balanced accuracy for multi-class classification, the concordance index (C-index) for survival, and quadratic weighted Cohen's $\kappa$ for tumor grading. Each task is run three times, and we report the mean over all data folds and random seeds using the provided data splits and task metadata.

For all benchmarks, results for both base and fine-tuned models were generated using the official implementations and default hyperparameters, with no additional tuning. Concatenation of \texttt{[CLS]} and mean-pooled patch features was performed for all models in PathoROB; only for Virchow2 in HEST and Patho-Bench; and for Virchow2, AquaViT, H0-mini and Midnight-12k in THUNDER, following the original implementations.

\section{Results}
\label{sec:results}

\subsection{Fine-tuning jointly improves robustness and performance}
\label{sec:results-joint}

Table~\ref{tab:robsummary_compact} (left) reports the PathoROB robustness index for each base and fine-tuned model pair. The robustness index increases consistently for every model ($p < 10^{-4}$, one-sided Wilcoxon signed-rank test), with some substantial improvements for Phikon-v2 ($0.47\rightarrow0.81$), Prov-GigaPath ($0.62\rightarrow0.89$) and H-Optimus-0 ($0.81\rightarrow0.92$).

Table~\ref{tab:robsummary_compact} (right) aggregates the performance on HEST, THUNDER and Patho-Bench into a single cross-benchmark ranking. Fine-tuned models dominate the leaderboard: the best fine-tuned encoder (UNI2-h) attains a total rank of $5$ (versus $21$ for the strongest base model), and every model improves its overall rank after fine-tuning ($p < 10^{-4}$, one-sided Wilcoxon signed rank test). Together with the robustness gains, this yields the joint up-and-to-the-right shift visualized in Figure~\ref{fig:abstract}: robustness and performance improve simultaneously, with no observed trade-off.

According to the complete leaderboards available in Supplementary \ref{app:leaderboards}, all top-1 positions are taken by fine-tuned models: UNI2-h for HEST (0.4290), UNI2-h for THUNDER (30 against 35 for the base UNI2-h), and Midnight-12k for Patho-Bench. On PathoROB, 8 out of the 10 most robust models are fine-tuned models.

\input{tables/inc/robsummary_compact_redacted}

\begin{figure}[t]
  \centering
  \makebox[\linewidth][c]{\includegraphics[width=\linewidth]{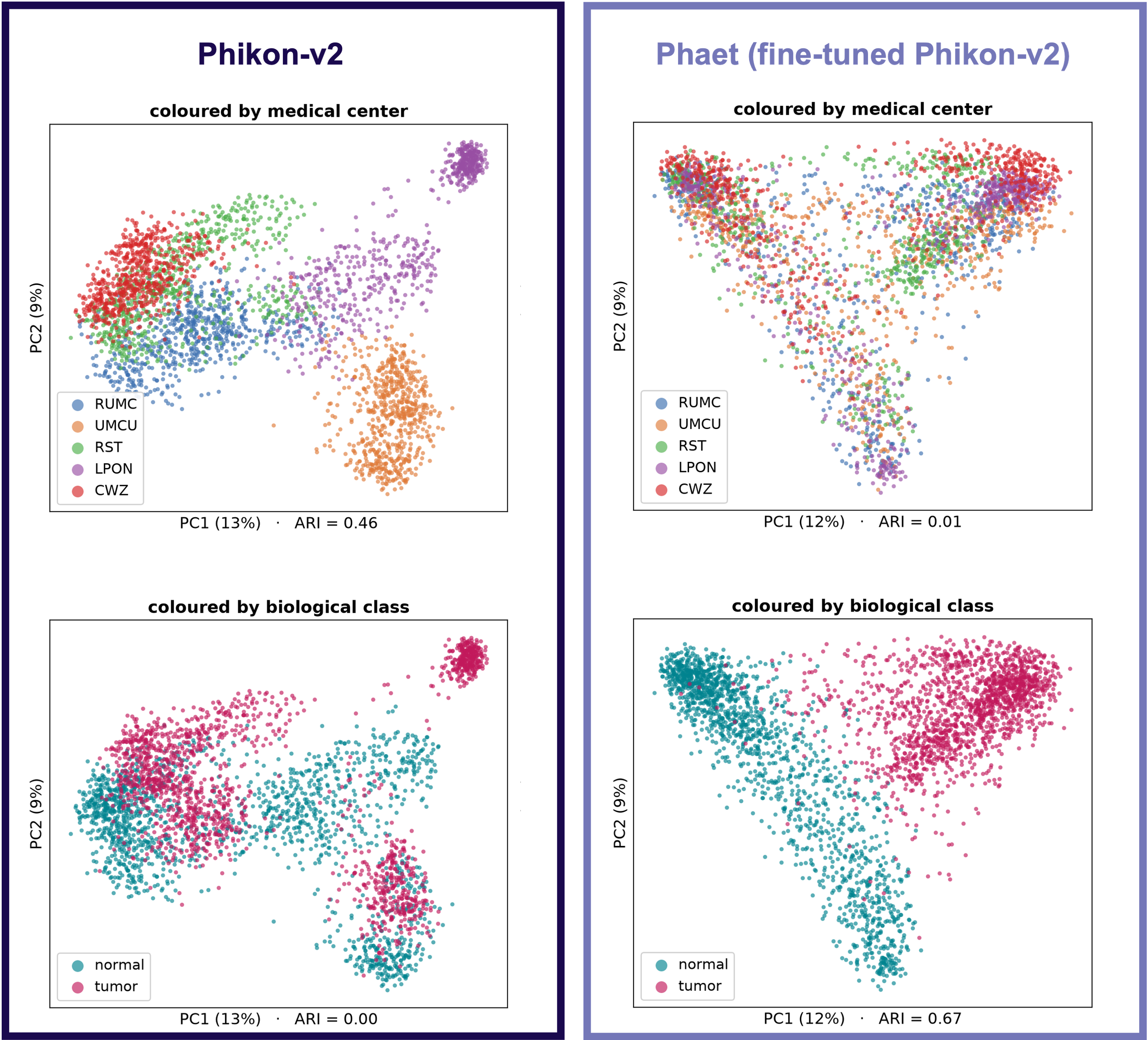}}
  \caption{\textbf{Fine-tuning reorganizes Phikon-v2's feature space around biology rather than
  acquisition site.} PCA of Phikon-v2 features on
  PathoROB-Camelyon dataset, before (left) and after (right) fine-tuning, colored by medical
  center (top) and by biological class (bottom). Each panel is annotated with the Adjusted Rand Index (ARI)
  between a silhouette-selected $k$-means clustering of the full-dimensional features and the corresponding
  labeling. Before fine-tuning, the representation clusters by medical center (ARI $0.46$ vs.\ $0.00$ for
  metastasis status); after fine-tuning, it clusters by biological class (ARI $0.67$ vs.\ $0.01$ for center).}
  \label{fig:phikon_refined}
\end{figure}

Figure~\ref{fig:phikon_refined} shows a two-dimensional PCA projection of Phikon-v2 features extracted for the PathoROB-Camelyon dataset~\citep{komen2026robust}. Tiles are sentinel lymph-node patches from breast-cancer patients drawn from the Camelyon16 and Camelyon17 cohorts across five Dutch medical centers (RUMC, UMCU, CWZ, RST, LPON), each labeled as containing nodal metastasis (tumor) or not (normal). The five centers were digitized on three distinct whole-slide scanners: RUMC, CWZ and RST slides on a 3DHistech P250 ($0.24\,\mu$m/px), UMCU on a Hamamatsu XR C12000 ($0.23\,\mu$m/px), and LPON on a Philips IntelliSite Ultra-Fast scanner ($0.25\,\mu$m/px). For the original Phikon-v2 model, the leading axis of variation separates the acquisition pipeline rather than the tissue (top left): a $k$-means partition of the full-dimensional features ($k \in \{1,...,10\}$ optimizes the silhouette score of the corresponding clustering solution) aligns with the medical center but not the biology (Adjusted Rand Index, ARI, of $0.46$ against center vs.\ $0.00$ against metastasis status). In particular, the three centers sharing the 3DHistech P250 (RUMC, CWZ, RST) collapse into a single cloud, whereas UMCU and LPON (each on a distinct scanner) form their own clusters. After fine-tuning, the centers become intermixed (top-right, ARI $0.01$) while metastatic and normal tiles separate (bottom-right, ARI $0.67$). Fine-tuning re-purposes the dominant directions of the feature space from encoding where a WSI was digitized to encoding what the morphology is. Importantly, none of these five centers were seen during fine-tuning, so the effect reflects generalization of the learned invariance to unseen acquisition sources.

\newpage

\subsection{Downstream performance}
\label{sec:results-performance}

A robustification recipe is only useful if it preserves the generic quality of the representation. Tables~\ref{tab:thunder_compact}, \ref{tab:hest_cls}, and \ref{tab:pathobench} report models evaluation across THUNDER, HEST, and Patho-Bench, respectively. Fine-tuned models match or exceed their base counterparts on the aggregate metrics, indicating that fine-tuning does not come at the expense of downstream utility.

\paragraph{THUNDER} Table~\ref{tab:thunder_compact} shows that fine-tuning lowers the rank sum for 8 of the 10 models, reducing the mean rank sum from $62$ to $49$ ($p=7.5\times10^{-3}$, one-sided Wilcoxon signed-rank test). The gains are largest for the models that start furthest behind: Midnight-12k ($70\rightarrow34$), H-Optimus-0 ($68\rightarrow41$) and Prov-GigaPath ($78\rightarrow52$); while the strongest base model, UNI2-h, remains on top and still improves ($26\rightarrow20$). Only H0-mini regresses slightly ($61\rightarrow65$) and AquaViT is unchanged, confirming that fine-tuning does not erode general tile-level representation quality.

\input{tables/inc/thunder_compact}

\newpage

\paragraph{HEST} For gene-expression prediction (Table~\ref{tab:hest_cls}), the average Pearson correlation increases for 9 of the 10 models ($p=2.9\times10^{-3}$, one-sided Wilcoxon signed-rank test) with the mean rising from $0.398$ to $0.411$; the sole exception, GenBio-PathFM, is essentially unchanged ($-0.002$). Improvements are consistent across the nine cancer types and largest for Prov-GigaPath ($+0.022$), Midnight-12k ($+0.022$) and Phikon-v2 ($+0.020$). After fine-tuning, H-Optimus-0 and UNI2-h jointly reach the best average correlation ($0.4290$).

\input{tables/inc/hest_cls}

\paragraph{Patho-Bench} On slide-level tasks, fine-tuning improves the overall average score for all ten models ($p=9.8\times10^{-4}$, one-sided Wilcoxon signed-rank) from $54.9$ to $56.6$ (Table~\ref{tab:pathobench}). Gains are spread across the molecular, morphological, survival and treatment-response groups, and are largest for Midnight-12k ($+3.4$), which becomes the best fine-tuned model overall ($58.0$), followed by H-Optimus-0 ($+2.5$) and Prov-GigaPath ($+2.2$). \\\\

\input{tables/inc/pathobench_redacted}

\newpage

\section{Understanding the robustness gains}
\label{sec:analysis}

To probe how acquisition factors are encoded in pathology FMs, we rely on two publicly available datasets that were built to isolate acquisition variability from tissue content. The PLISM dataset~\citep{ochi2024plism} is a group of consecutive slides digitized on 7 different scanners and stained across 13 H\&E conditions, so that each of the 91 resulting acquisition variants captures the same biological information, that is, a collection of 46 TMAs (Tissue Micro Arrays) from 46 different organs. All WSIs were spatially registered to a common reference slide (AT2 scanner, GIVH stain) with the Elastix~\citep{staring2010elastix} software and tessellated into 16,278 aligned tiles, so that every tile location has a matched counterpart in each variant. SCORPION~\citep{ryu2025scorpion} similarly provides 480 tissue samples each digitized on 5 scanners (2{,}400 spatially aligned patches across Leica Aperio AT2, Leica Aperio GT450, Roche Ventana DP200, 3DHistech P1000, and Philips UFS B300). 

\begin{figure}[h!]
  \centering
  \includegraphics[width=0.8\linewidth]{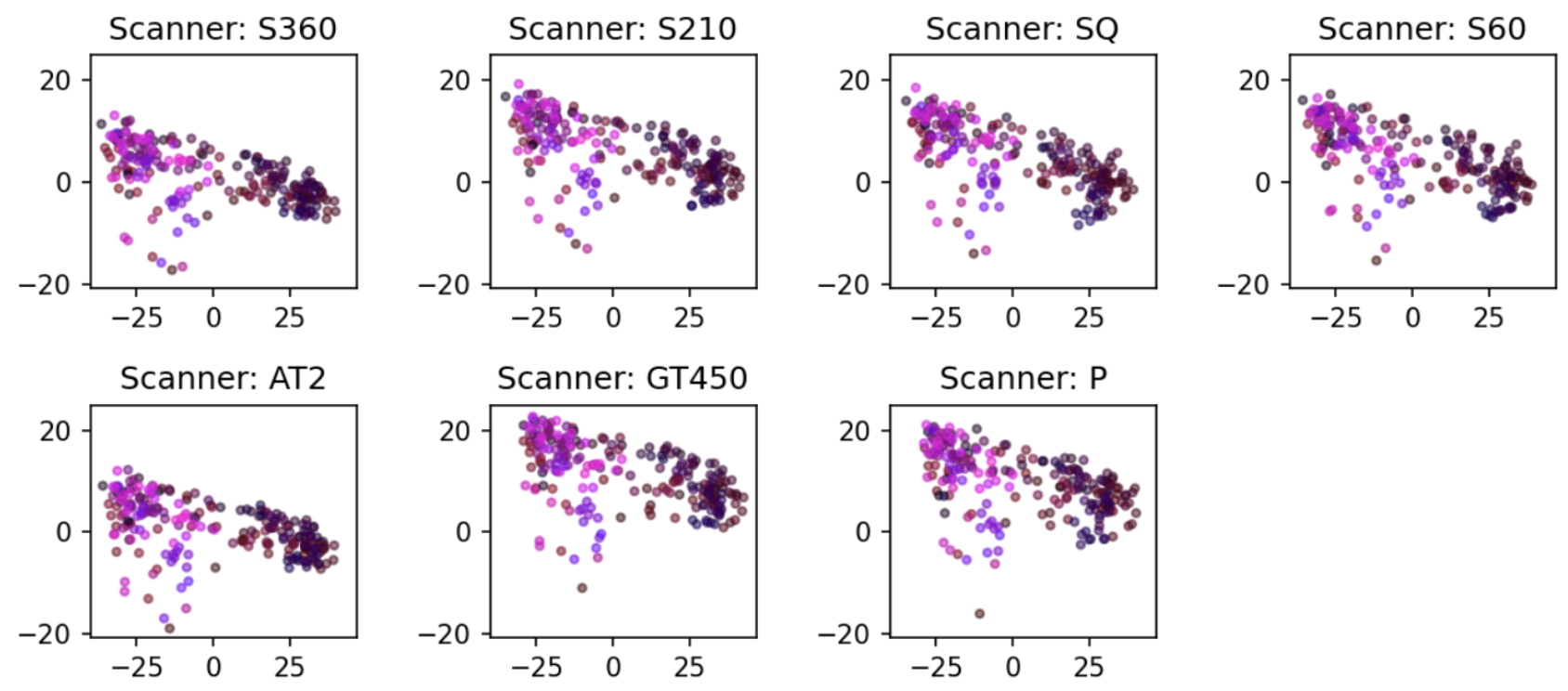}
  \caption{\textbf{Acquisition factors seem linearly encoded in feature space.} PCA projection of
  H-Optimus-0 tile features on the PLISM dataset for a fixed staining protocol (GIVH), one panel
  per scanner. Each of the 225 points is a colorectal-cancer (CRC) tile (a subset of the $16{,}278$
  tiles of the slide), colored by its spatial position within the slide. The point clouds keep a similar shape across scanners, suggesting that the scanner shift is a simple, near-linear offset in feature space.}
  \label{fig:scanner_variability}
\end{figure}

\begin{figure}[t]
  \centering
  \includegraphics[width=\linewidth]{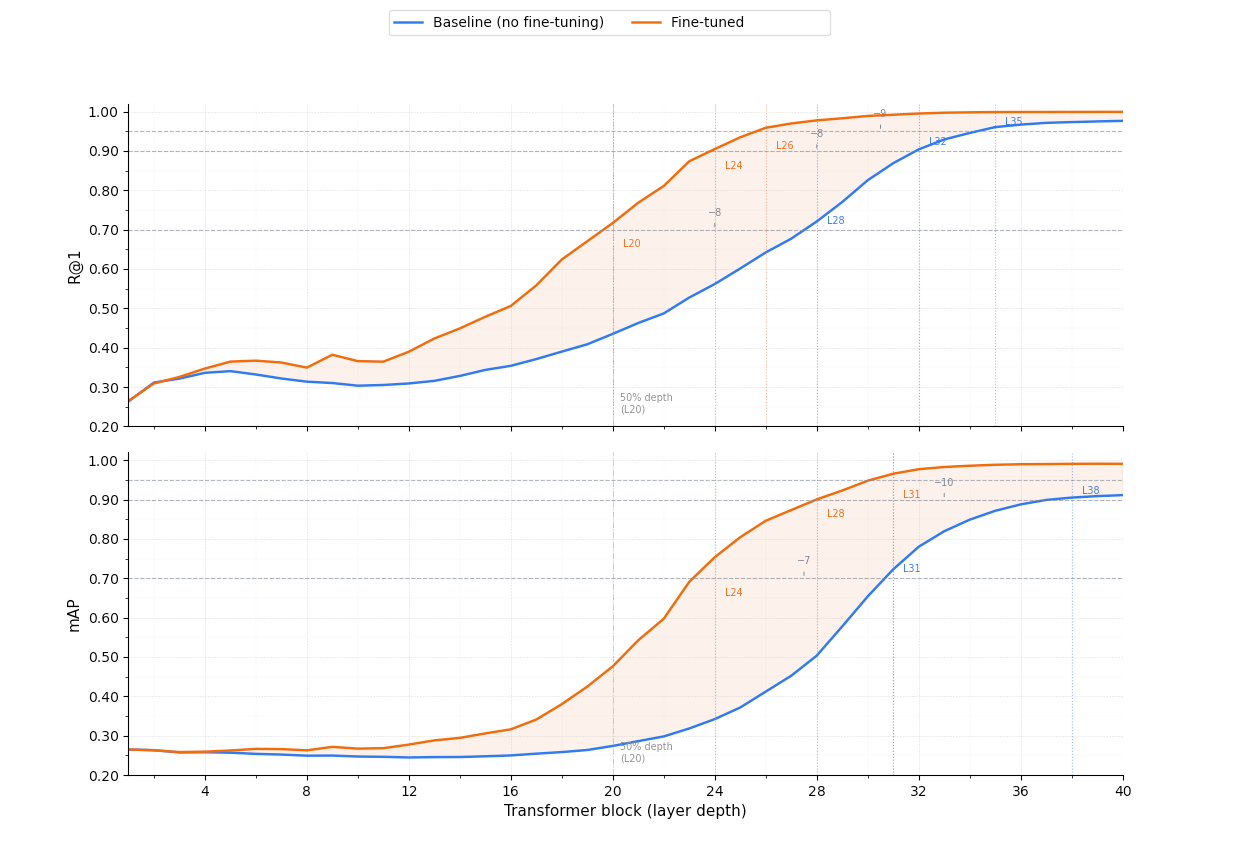}
  \caption{\textbf{Fine-tuning makes scanner invariance emerge earlier and stronger with depth.}
  Cross-scanner retrieval on SCORPION (R@1, top; mAP, bottom) as a function of the transformer block
  used to extract features, for the base H-Optimus-0 model (blue) and after fine-tuning (orange).
  Fine-tuning shifts the curves up and to the left, reaching comparable retrieval roughly eight blocks
  earlier and a higher final value.}
  \label{fig:scorpion_depth}
\end{figure}

To understand why fine-tuning yields such consistent improvements, we examine how acquisition factors are encoded in the feature space. Figure~\ref{fig:scanner_variability} suggests that the scanner shift is a near-linear offset in the output feature space that could be corrected through feature-level correction. A closer analysis (Figure~\ref{fig:scorpion_depth}) across network depth actually reveals that invariance is learned deep in the transformer blocks.

\paragraph{Acquisition factors seem linearly encoded in feature space.} Figure~\ref{fig:scanner_variability} projects tile features of the PLISM dataset to two dimensions, with one panel per scanner, obtained by PCA on H-Optimus-0 features for a fixed staining protocol (GIVH). Interestingly, this figure suggests that the scanner shift is a simple, near-linear offset in feature space: the per-scanner point clouds keep a similar shape (color spatial distribution is similar across scanners) and appear largely translated, which would imply that a feature-level correction could undo it.

\paragraph{Correcting acquisition shifts requires deep adaptation of the backbone.} Instead, we show how invariance builds up across the transformer depth using cross-scanner retrieval on SCORPION~\citep{ryu2025scorpion}, built to isolate scanner-induced variability from tissue content. Given a query patch imaged on one scanner, the task is to retrieve the same physical patch captured by the other scanners; we report Recall@1 (R@1) and mean average precision (mAP). To make the task discriminative, each $1024\times1024$ patch is partitioned into 16 tiles ($256\times256$, resized to $224\times224$), and the negatives for a query include the other tiles of its own grid as well as all tiles of the other slides. Because adjacent grid tiles cover neighboring tissue, they are visually near-identical, so a scanner-sensitive encoder easily ranks such a spatial look-alike (or a same-scanner copy) above the true cross-scanner match. Attaining high R@1 and mAP therefore requires a representation that is at once invariant to the scanner and discriminative of fine spatial detail.

\newpage

Figure~\ref{fig:scorpion_depth} reports R@1 and mAP as a function of the transformer block from which tile features are extracted, for the base (blue) and fine-tuned (orange) H-Optimus-0 encoder. For the base model, robust cross-scanner matching only develops in the last few blocks; fine-tuning shifts the entire curve up and to the left, reaching a given retrieval quality roughly eight blocks earlier and attaining a higher asymptote (mAP $\approx 0.99$ vs $0.91$ at the final block). Fine-tuning therefore does not merely re-tune the output layer: it instills scanner invariance progressively throughout the network, so that even intermediate representations become markedly more robust. Taken together, these observations suggest that purely feature-level operations (e.g., statistics matching or CORAL-style covariance alignment) can partially reduce the shift but are not deep enough to remove it entirely.

\section{Conclusion}\label{sec:conclusion}

In this work, we show that fine-tuning can improve robustness and downstream performance simultaneously (Figure~\ref{fig:abstract}). Enforcing invariance to acquisition factors could plausibly erase biologically useful signal and degrade downstream tasks. Instead, every fine-tuned model moves up and to the right, suggesting that scanner and stain-related directions in the feature space are largely nuisance dimensions; removing them frees capacity for biologically relevant structure rather than competing with it. This interpretation is consistent with prior evidence that acquisition confounders are strongly encoded by current FMs~\citep{dejong2025unrobust, komen2026robust}.

The gains are largest for the least robust base models: Phikon-v2 ($0.47\rightarrow0.81$ robustness index) and Prov-GigaPath ($0.62\rightarrow0.89$), while already-strong encoders such as UNI2-h and Virchow2 improve more modestly yet remain at the top. Fine-tuning therefore acts as an equalizer: a lightweight, label-free fine-tuning step that narrows the gap between encoders trained at different scales with different pretraining data diversity.

We hypothesize that the joint improvement of robustness and performance comes from the fact that confounder-related directions act as structured noise for downstream predictors: if acquisition signal is entangled in the features, a classifier will tend to learn spurious shortcuts that prevent generalizability. By collapsing these directions, fine-tuning removes this nuisance variance and leaves a cleaner, lower-noise representation in which the biologically relevant signal is easier to separate.

In terms of limitations, some fine-tuned models show slightly degraded performance compared to their base counterparts (e.g., H0-mini on THUNDER; GenBio-PathFM on HEST). Lastly, our evaluation is limited to vision-only encoders; the effect of our fine-tuning approach on vision-language or multi-modal encoders remains to be explored. Further, we measure robustness through the PathoROB index, which quantifies the degree to which biological features dominate over non-biological confounders in the feature space. While this is a useful proxy, it does not directly measure downstream robustness under domain shift for specific tasks such as biomarker prediction or survival analysis~\citep{scanner2026thiringer, schonpflug2026protocol}. Evaluating on a broader set of robustness benchmarks would further clarify the generality of the approach.

\begin{ack}

\paragraph{Computing resources.} This work was granted access to the High-Performance
Computing (HPC) resources of IDRIS under the allocation 2026-A0201012519 made by GENCI. Fine-tuning experiments were performed using the EuroHPC supercomputer MareNostrum 5, hosted by the Barcelona Supercomputing Center (BSC). We gratefully acknowledge EuroHPC and BSC for providing access to these resources.

\paragraph{Data access.} The results presented here are in part based upon data generated
by the TCGA Research Network: https://www.cancer.gov/tcga.

\end{ack}

\small
\bibliographystyle{plainnat}
\bibliography{references}

\newpage
\appendix

\section{Extended leaderboards}
\label{app:leaderboards}

The following tables report the complete leaderboards for all four benchmarks. For previously released models we quote the official published values, whereas results for our fine-tuned encoders were computed in-house under the same evaluation protocols.

\input{tables/inc/thunder_full_redacted}

\input{tables/inc/hest_cls_full_redacted}

\input{tables/inc/pathorob_full_redacted}



\end{document}

%% file: tables/inc/robsummary_compact_redacted.tex
\begin{center}
\centering
\small
\captionof{table}{Robustness and cross-benchmark performance summary. \textbf{Left} (PathoROB): robustness index (RI) per dataset and average~($\uparrow$). \textbf{Right}: rank among 20 models (1\,=\,best) on each benchmark with the average metric in parentheses (HEST: average Pearson~$\uparrow$, THUNDER: rank sum~$\downarrow$, Patho-bench: grand average~$\uparrow$), and total being the sum of the three benchmark ranks~$\downarrow$. \textbf{Bold} marks the better value within each pair. Colors: {\setlength{\fboxsep}{1pt}\colorbox{green!35}{green}}\,=\,improvement of the fine-tuned model over the base model, {\setlength{\fboxsep}{1pt}\colorbox{red!35}{red}}\,=\,regression, {\setlength{\fboxsep}{1pt}\colorbox{orange!50}{orange}}\,=\,tie.}
\label{tab:robsummary_compact}
\resizebox{\textwidth}{!}{%
\begin{tabular}{@{}ll rrrr | rrrr@{}}
\toprule
\textbf{Model} & \textbf{Variant} & \multicolumn{4}{c}{\textbf{Robustness Index (PathoROB)}} & \multicolumn{4}{|c}{\textbf{Performance (ranks~$\downarrow$)}} \\[2pt]
\cmidrule(lr){3-6}\cmidrule(lr){7-10}
 & & \multicolumn{1}{c}{{\strut TCGA~RI~$\uparrow$}} & \multicolumn{1}{c}{{\strut Cam.~RI~$\uparrow$}} & \multicolumn{1}{c}{{\strut Tolkach~RI~$\uparrow$}} & \textbf{Average} & \multicolumn{1}{c}{{\strut HEST~$\uparrow$}} & \multicolumn{1}{c}{{\strut THUNDER~$\downarrow$}} & \multicolumn{1}{c}{{\strut Patho-bench~$\uparrow$}} & \multicolumn{1}{c}{{\strut Total~$\downarrow$}} \\
\midrule
UNI2-h & Base & 0.803 & 0.544 & 0.923 & 0.757 & 7~(0.4141) & 2~(26) & 12~(55.6) & 21 \\
\rowcolor{shadegray} & \textit{Fine-tuned} & \textbf{0.863} & \textbf{0.901} & \textbf{0.960} & \cellcolor{green!35}\textbf{0.908} & \cellcolor{green!35}\textbf{2~(0.4290)} & \cellcolor{green!35}\textbf{1~(20)} & \cellcolor{green!35}\textbf{2~(57.3)} & \cellcolor{green!35}\textbf{5} \\
\addlinespace[2pt]
\midrule
Midnight-12k & Base & 0.858 & 0.478 & 0.941 & 0.759 & 15~(0.3952) & 15~(70) & 15~(54.6) & 45 \\
\rowcolor{shadegray} \textit{Mascaret} & \textit{Fine-tuned} & \textbf{0.893} & \textbf{0.907} & \textbf{0.972} & \cellcolor{green!35}\textbf{0.924} & \cellcolor{green!35}\textbf{5~(0.4167)} & \cellcolor{green!35}\textbf{5~(34)} & \cellcolor{green!35}\textbf{1~(58.0)} & \cellcolor{green!35}\textbf{11} \\
\addlinespace[2pt]
\midrule
H-Optimus-0 & Base & 0.812 & 0.703 & 0.918 & 0.811 & 6~(0.4150) & 14~(68) & 17~(54.4) & 37 \\
\rowcolor{shadegray} & \textit{Fine-tuned} & \textbf{0.856} & \textbf{0.933} & \textbf{0.961} & \cellcolor{green!35}\textbf{0.917} & \cellcolor{green!35}\textbf{1~(0.4290)} & \cellcolor{green!35}\textbf{7~(41)} & \cellcolor{green!35}\textbf{5~(56.8)} & \cellcolor{green!35}\textbf{13} \\
\addlinespace[2pt]
\midrule
GenBio-PathFM & Base & 0.838 & 0.862 & 0.959 & 0.886 & \textbf{3~(0.4197)} & 6~(40) & 19~(54.0) & 28 \\
\rowcolor{shadegray} & \textit{Fine-tuned} & \textbf{0.863} & \textbf{0.926} & \textbf{0.966} & \cellcolor{green!35}\textbf{0.918} & \cellcolor{red!35}4~(0.4178) & \cellcolor{green!35}\textbf{3~(29)} & \cellcolor{green!35}\textbf{11~(55.8)} & \cellcolor{green!35}\textbf{18} \\
\addlinespace[2pt]
\midrule
Virchow2 & Base & 0.822 & 0.799 & 0.954 & 0.858 & 13~(0.4034) & 8~(43) & 13~(55.4) & 34 \\
\rowcolor{shadegray} & \textit{Fine-tuned} & \textbf{0.849} & \textbf{0.935} & \textbf{0.969} & \cellcolor{green!35}\textbf{0.918} & \cellcolor{green!35}\textbf{8~(0.4135)} & \cellcolor{green!35}\textbf{4~(32)} & \cellcolor{green!35}\textbf{6~(56.8)} & \cellcolor{green!35}\textbf{18} \\
\addlinespace[2pt]
\midrule
AquaViT & Base & 0.781 & 0.673 & 0.925 & 0.793 & 12~(0.4045) & 10~(60) & 9~(56.2) & 31 \\
\rowcolor{shadegray} & \textit{Fine-tuned} & \textbf{0.811} & \textbf{0.865} & \textbf{0.950} & \cellcolor{green!35}\textbf{0.875} & \cellcolor{green!35}\textbf{11~(0.4064)} & \cellcolor{orange!50}10~(60) & \cellcolor{green!35}\textbf{3~(57.1)} & \cellcolor{green!35}\textbf{24} \\
\addlinespace[2pt]
\midrule
Prov-GigaPath & Base & 0.737 & 0.382 & 0.746 & 0.622 & 17~(0.3875) & 16~(78) & 16~(54.5) & 49 \\
\rowcolor{shadegray} & \textit{Fine-tuned} & \textbf{0.827} & \textbf{0.872} & \textbf{0.956} & \cellcolor{green!35}\textbf{0.885} & \cellcolor{green!35}\textbf{9~(0.4098)} & \cellcolor{green!35}\textbf{9~(52)} & \cellcolor{green!35}\textbf{8~(56.7)} & \cellcolor{green!35}\textbf{26} \\
\addlinespace[2pt]
\midrule
H0-mini & Base & 0.794 & 0.717 & 0.932 & 0.814 & 14~(0.3958) & \textbf{12~(61)} & 7~(56.7) & 33 \\
\rowcolor{shadegray} & \textit{Fine-tuned} & \textbf{0.811} & \textbf{0.842} & \textbf{0.950} & \cellcolor{green!35}\textbf{0.868} & \cellcolor{green!35}\textbf{10~(0.4076)} & \cellcolor{red!35}13~(65) & \cellcolor{green!35}\textbf{4~(56.9)} & \cellcolor{green!35}\textbf{27} \\
\addlinespace[2pt]
\midrule
Phikon-v2 & Base & 0.619 & 0.019 & 0.768 & 0.469 & 19~(0.3747) & 20~(97) & 18~(54.1) & 57 \\
\rowcolor{shadegray} \textit{Phaet} & \textit{Fine-tuned} & \textbf{0.785} & \textbf{0.702} & \textbf{0.932} & \cellcolor{green!35}\textbf{0.806} & \cellcolor{green!35}\textbf{16~(0.3943)} & \cellcolor{green!35}\textbf{18~(83)} & \cellcolor{green!35}\textbf{10~(55.8)} & \cellcolor{green!35}\textbf{44} \\
\addlinespace[2pt]
\midrule
Phikon & Base & 0.623 & 0.011 & 0.795 & 0.476 & 20~(0.3660) & 18~(83) & 20~(53.3) & 58 \\
\rowcolor{shadegray} & \textit{Fine-tuned} & \textbf{0.731} & \textbf{0.244} & \textbf{0.914} & \cellcolor{green!35}\textbf{0.630} & \cellcolor{green!35}\textbf{18~(0.3832)} & \cellcolor{green!35}\textbf{17~(81)} & \cellcolor{green!35}\textbf{14~(55.0)} & \cellcolor{green!35}\textbf{49} \\
\bottomrule
\end{tabular}}
\end{center}

%% file: tables/inc/thunder_compact.tex
\begin{center}
\centering
\small
\captionof{table}{THUNDER benchmark. Per-task scores with leaderboard rank in parentheses; rank sum~$\downarrow$ is the primary metric. \textbf{Bold} marks the better value within each pair. Colors: {\setlength{\fboxsep}{1pt}\colorbox{green!35}{green}}\,=\,improvement of the fine-tuned model over the base model, {\setlength{\fboxsep}{1pt}\colorbox{red!35}{red}}\,=\,regression, {\setlength{\fboxsep}{1pt}\colorbox{orange!50}{orange}}\,=\,tie.}
\label{tab:thunder_compact}
\resizebox{\textwidth}{!}{%
\begin{tabular}{@{}llrrrrrrr@{}}
\toprule
\textbf{Model} & \textbf{Variant} & \multicolumn{1}{c}{{\strut KNN~$\uparrow$}} & \multicolumn{1}{c}{{\strut Lin.~prob.~$\uparrow$}} & \multicolumn{1}{c}{{\strut Few-shot~$\uparrow$}} & \multicolumn{1}{c}{{\strut Seg.~$\uparrow$}} & \multicolumn{1}{c}{{\strut Calib.~$\downarrow$}} & \multicolumn{1}{c}{{\strut Adv.~att.~$\downarrow$}} & \multicolumn{1}{c}{{\strut Rank~sum~$\downarrow$}} \\
\midrule
UNI2-h & Base & 83.3~(4) & \textbf{86.3~(1)} & \textbf{79.8~(1)} & \textbf{68.1~(3)} & 3.7~(8) & 31.0~(9) & 26 \\
\rowcolor{tablegray} & \textit{Fine-tuned} & \textbf{83.4~(3)} & 85.5~(2) & 79.5~(3) & 67.6~(6) & \textbf{2.5~(3)} & \textbf{24.1~(3)} & \cellcolor{green!35}\textbf{20} \\
\addlinespace[2pt]
\midrule
GenBio-PathFM & Base & 83.5~(2) & 85.1~(4) & 79.4~(4) & \textbf{67.2~(8)} & \textbf{3.7~(8)} & 32.7~(14) & 40 \\
\rowcolor{tablegray} & \textit{Fine-tuned} & \textbf{83.9~(1)} & \textbf{85.3~(3)} & \textbf{79.6~(2)} & 66.8~(9) & 4.0~(9) & \textbf{26.2~(5)} & \cellcolor{green!35}\textbf{29} \\
\addlinespace[2pt]
\midrule
Virchow2 & Base & \textbf{82.9~(5)} & 84.8~(5) & 73.9~(14) & \textbf{68.2~(2)} & \textbf{3.6~(7)} & 31.1~(10) & 43 \\
\rowcolor{tablegray} & \textit{Fine-tuned} & 82.6~(6) & \textbf{85.1~(4)} & \textbf{76.6~(7)} & 68.0~(4) & 4.2~(10) & \textbf{7.7~(1)} & \cellcolor{green!35}\textbf{32} \\
\addlinespace[2pt]
\midrule
Midnight-12k & Base & 80.0~(13) & 84.4~(7) & 71.5~(20) & 66.0~(12) & 2.4~(2) & 35.7~(16) & 70 \\
\rowcolor{tablegray} \textit{Mascaret} & \textit{Fine-tuned} & \textbf{81.7~(8)} & \textbf{84.6~(6)} & \textbf{75.2~(11)} & \textbf{67.6~(6)} & \textbf{2.3~(1)} & \textbf{23.2~(2)} & \cellcolor{green!35}\textbf{34} \\
\addlinespace[2pt]
\midrule
H-Optimus-0 & Base & 81.5~(9) & 83.7~(9) & 76.2~(8) & 63.5~(15) & 3.6~(7) & 42.1~(20) & 68 \\
\rowcolor{tablegray} & \textit{Fine-tuned} & \textbf{81.9~(7)} & \textbf{84.0~(8)} & \textbf{77.4~(5)} & \textbf{68.1~(3)} & \textbf{3.2~(5)} & \textbf{32.4~(13)} & \cellcolor{green!35}\textbf{41} \\
\addlinespace[2pt]
\midrule
Prov-GigaPath & Base & 79.4~(15) & 82.4~(13) & 75.5~(10) & 61.1~(16) & 3.4~(6) & 40.6~(18) & 78 \\
\rowcolor{tablegray} & \textit{Fine-tuned} & \textbf{80.8~(10)} & \textbf{83.0~(11)} & \textbf{76.7~(6)} & \textbf{65.8~(13)} & \textbf{3.2~(5)} & \textbf{28.6~(7)} & \cellcolor{green!35}\textbf{52} \\
\addlinespace[2pt]
\midrule
AquaViT & Base & \textbf{80.5~(11)} & \textbf{82.7~(12)} & \textbf{75.7~(9)} & 68.0~(4) & \textbf{5.2~(13)} & 32.0~(11) & 60 \\
\rowcolor{tablegray} & \textit{Fine-tuned} & 80.1~(12) & 82.4~(13) & 75.1~(12) & \textbf{68.3~(1)} & 5.3~(14) & \textbf{29.3~(8)} & \cellcolor{orange!50}60 \\
\addlinespace[2pt]
\midrule
H0-mini & Base & \textbf{79.7~(14)} & \textbf{83.5~(10)} & \textbf{75.0~(13)} & \textbf{67.7~(5)} & \textbf{3.0~(4)} & 33.7~(15) & \textbf{61} \\
\rowcolor{tablegray} & \textit{Fine-tuned} & 78.7~(16) & 82.4~(13) & 73.8~(15) & 67.5~(7) & 3.7~(8) & \textbf{27.6~(6)} & \cellcolor{red!35}65 \\
\addlinespace[2pt]
\midrule
Phikon & Base & 75.7~(19) & \textbf{80.2~(15)} & \textbf{73.6~(16)} & \textbf{66.8~(9)} & \textbf{5.1~(12)} & 32.2~(12) & 83 \\
\rowcolor{tablegray} & \textit{Fine-tuned} & \textbf{76.1~(18)} & 79.8~(16) & 73.2~(18) & 66.6~(10) & 5.4~(15) & \textbf{24.5~(4)} & \cellcolor{green!35}\textbf{81} \\
\addlinespace[2pt]
\midrule
Phikon-v2 & Base & 74.0~(20) & 79.3~(17) & 71.8~(19) & \textbf{66.5~(11)} & 4.5~(11) & 41.9~(19) & 97 \\
\rowcolor{tablegray} \textit{Phaet} & \textit{Fine-tuned} & \textbf{77.7~(17)} & \textbf{80.7~(14)} & \textbf{73.3~(17)} & 65.3~(14) & \textbf{3.0~(4)} & \textbf{38.8~(17)} & \cellcolor{green!35}\textbf{83} \\
\bottomrule
\end{tabular}}
\end{center}

%% file: tables/inc/hest_cls.tex
\begin{center}
\centering
\small
\captionof{table}{HEST benchmark. We report Pearson correlation~($\uparrow$) between predicted and measured gene expression across 9 cancer types. \textbf{Bold} marks the better value within each model pair. Colors: {\setlength{\fboxsep}{1pt}\colorbox{green!35}{green}}\,=\,improvement of the fine-tuned model over the base model, {\setlength{\fboxsep}{1pt}\colorbox{red!35}{red}}\,=\,regression.}
\label{tab:hest_cls}
\resizebox{\textwidth}{!}{%
\begin{tabular}{@{}llrrrrrrrrrr@{}}
\toprule
\textbf{Model} & \textbf{Variant} & \multicolumn{1}{c}{{\strut IDC}} & \multicolumn{1}{c}{{\strut PRAD}} & \multicolumn{1}{c}{{\strut PAAD}} & \multicolumn{1}{c}{{\strut SKCM}} & \multicolumn{1}{c}{{\strut COAD}} & \multicolumn{1}{c}{{\strut READ}} & \multicolumn{1}{c}{{\strut CCRCC}} & \multicolumn{1}{c}{{\strut LUNG}} & \multicolumn{1}{c}{{\strut LYMPH$_{\text{IDC}}$}} & \textbf{Average} \\
\midrule
H-Optimus-0 & Base & 0.5976 & \textbf{0.3848} & 0.4911 & 0.6454 & 0.3086 & 0.2216 & 0.2676 & 0.5590 & 0.2591 & 0.4150 \\
\rowcolor{tablegray} & \textit{Fine-tuned} & \textbf{0.6069} & 0.3695 & \textbf{0.5205} & \textbf{0.6764} & \textbf{0.3136} & \textbf{0.2466} & \textbf{0.2791} & \textbf{0.5811} & \textbf{0.2671} & \cellcolor{green!35}\textbf{0.4290} \\
\addlinespace[2pt]
\midrule
UNI2-h & Base & 0.5898 & 0.3569 & 0.5001 & 0.6606 & 0.3015 & 0.2223 & 0.2640 & 0.5587 & 0.2727 & 0.4141 \\
\rowcolor{tablegray} & \textit{Fine-tuned} & \textbf{0.6016} & \textbf{0.3763} & \textbf{0.5290} & \textbf{0.6781} & \textbf{0.3294} & \textbf{0.2239} & \textbf{0.2734} & \textbf{0.5729} & \textbf{0.2761} & \cellcolor{green!35}\textbf{0.4290} \\
\addlinespace[2pt]
\midrule
GenBio-PathFM & Base & \textbf{0.5872} & 0.3913 & 0.4959 & \textbf{0.6715} & \textbf{0.3284} & 0.1785 & \textbf{0.2615} & \textbf{0.5787} & \textbf{0.2842} & \textbf{0.4197} \\
\rowcolor{tablegray} & \textit{Fine-tuned} & 0.5867 & \textbf{0.3965} & \textbf{0.5066} & 0.6473 & 0.3217 & \textbf{0.1961} & 0.2436 & 0.5783 & 0.2831 & \cellcolor{red!35}0.4178 \\
\addlinespace[2pt]
\midrule
Midnight-12k & Base & 0.5823 & 0.3370 & 0.4900 & 0.6360 & 0.2908 & 0.1856 & 0.2132 & 0.5577 & 0.2642 & 0.3952 \\
\rowcolor{tablegray} \textit{Mascaret} & \textit{Fine-tuned} & \textbf{0.5920} & \textbf{0.3760} & \textbf{0.5127} & \textbf{0.6442} & \textbf{0.3322} & \textbf{0.1964} & \textbf{0.2435} & \textbf{0.5802} & \textbf{0.2727} & \cellcolor{green!35}\textbf{0.4167} \\
\addlinespace[2pt]
\midrule
Virchow2 & Base & \textbf{0.5971} & 0.3529 & 0.4779 & 0.6402 & 0.2581 & \textbf{0.2074} & \textbf{0.2719} & \textbf{0.5685} & 0.2568 & 0.4034 \\
\rowcolor{tablegray} & \textit{Fine-tuned} & 0.5867 & \textbf{0.3900} & \textbf{0.5005} & \textbf{0.6543} & \textbf{0.2940} & 0.2056 & 0.2652 & 0.5656 & \textbf{0.2597} & \cellcolor{green!35}\textbf{0.4135} \\
\addlinespace[2pt]
\midrule
Prov-GigaPath & Base & 0.5515 & 0.3699 & 0.4746 & 0.5619 & 0.2992 & \textbf{0.1961} & 0.2430 & 0.5412 & 0.2500 & 0.3875 \\
\rowcolor{tablegray} & \textit{Fine-tuned} & \textbf{0.5856} & \textbf{0.3730} & \textbf{0.5138} & \textbf{0.6137} & \textbf{0.3214} & 0.1891 & \textbf{0.2607} & \textbf{0.5663} & \textbf{0.2648} & \cellcolor{green!35}\textbf{0.4098} \\
\addlinespace[2pt]
\midrule
H0-mini & Base & \textbf{0.5862} & 0.3687 & 0.4919 & 0.6012 & 0.2494 & 0.1863 & 0.2670 & 0.5482 & \textbf{0.2629} & 0.3958 \\
\rowcolor{tablegray} & \textit{Fine-tuned} & 0.5820 & \textbf{0.3805} & \textbf{0.5070} & \textbf{0.6305} & \textbf{0.2838} & \textbf{0.1874} & \textbf{0.2681} & \textbf{0.5662} & 0.2626 & \cellcolor{green!35}\textbf{0.4076} \\
\addlinespace[2pt]
\midrule
AquaViT & Base & 0.5875 & \textbf{0.3814} & 0.4763 & \textbf{0.6294} & \textbf{0.2984} & \textbf{0.2200} & 0.2317 & 0.5510 & \textbf{0.2649} & 0.4045 \\
\rowcolor{tablegray} & \textit{Fine-tuned} & \textbf{0.5928} & 0.3738 & \textbf{0.5020} & 0.6232 & 0.2935 & 0.2158 & \textbf{0.2378} & \textbf{0.5550} & 0.2635 & \cellcolor{green!35}\textbf{0.4064} \\
\addlinespace[2pt]
\midrule
Phikon-v2 & Base & 0.5408 & 0.3545 & 0.4455 & 0.5554 & 0.2500 & \textbf{0.1749} & 0.2659 & 0.5419 & 0.2437 & 0.3747 \\
\rowcolor{tablegray} \textit{Phaet} & \textit{Fine-tuned} & \textbf{0.5630} & \textbf{0.3546} & \textbf{0.4748} & \textbf{0.5985} & \textbf{0.2915} & 0.1696 & \textbf{0.2696} & \textbf{0.5622} & \textbf{0.2649} & \cellcolor{green!35}\textbf{0.3943} \\
\addlinespace[2pt]
\midrule
Phikon & Base & 0.5327 & 0.3420 & 0.4425 & 0.5355 & 0.2623 & 0.1532 & 0.2423 & 0.5466 & 0.2373 & 0.3660 \\
\rowcolor{tablegray} & \textit{Fine-tuned} & \textbf{0.5571} & \textbf{0.3639} & \textbf{0.4769} & \textbf{0.5615} & \textbf{0.2708} & \textbf{0.1597} & \textbf{0.2430} & \textbf{0.5649} & \textbf{0.2513} & \cellcolor{green!35}\textbf{0.3832} \\
\bottomrule
\end{tabular}}
\end{center}

%% file: tables/inc/pathobench_redacted.tex
{\fontsize{6.5}{8}\selectfont
\setlength{\tabcolsep}{1.2pt}
\begin{longtable}{@{}l*{10}{r >{\columncolor{shadegray}}r}@{}}
\caption{Patho-Bench benchmark. Scores are percentages (\%); metrics: AUC\,=\,Macro OvR AUC, bAcc\,=\,Balanced Accuracy, $\kappa$\,=\,Weighted Cohen kappa, C-idx\,=\,C-Index. Shaded columns (\textbf{model}$^\dagger$) report fine-tuned results; pairs ordered by fine-tuned grand average (descending). \textbf{Bold}: best score per row. Colors: {\setlength{\fboxsep}{1pt}\colorbox{green!35}{green}}\,=\,improvement of the fine-tuned model over the base model, {\setlength{\fboxsep}{1pt}\colorbox{red!35}{red}}\,=\,regression.}\label{tab:pathobench} \\
\toprule
\textbf{[Dataset][Task][Metric]} & \multicolumn{1}{c}{\rotatebox{90}{\strut Midnight-12k}} & \multicolumn{1}{>{\columncolor{shadegray}}c}{
\rotatebox{90}{\strut \textit{Mascaret}}} & \multicolumn{1}{c}{\rotatebox{90}{\strut UNI2-h}} & \multicolumn{1}{>{\columncolor{shadegray}}c}{
\rotatebox{90}{\strut UNI2-h$^\dagger$}} & \multicolumn{1}{c}{\rotatebox{90}{\strut AquaViT}} & \multicolumn{1}{>{\columncolor{shadegray}}c}{
\rotatebox{90}{\strut AquaViT$^\dagger$}} & \multicolumn{1}{c}{\rotatebox{90}{\strut H0-mini}} & \multicolumn{1}{>{\columncolor{shadegray}}c}{
\rotatebox{90}{\strut H0-mini$^\dagger$}} & \multicolumn{1}{c}{\rotatebox{90}{\strut H-Optimus-0}} & \multicolumn{1}{>{\columncolor{shadegray}}c}{
\rotatebox{90}{\strut H-Optimus-0$^\dagger$}} & \multicolumn{1}{c}{\rotatebox{90}{\strut Virchow2}} & \multicolumn{1}{>{\columncolor{shadegray}}c}{
\rotatebox{90}{\strut Virchow2$^\dagger$}} & \multicolumn{1}{c}{\rotatebox{90}{\strut Prov-GigaPath}} & \multicolumn{1}{>{\columncolor{shadegray}}c}{
\rotatebox{90}{\strut Prov-GigaPath$^\dagger$}} & \multicolumn{1}{c}{\rotatebox{90}{\strut Phikon-v2}} & \multicolumn{1}{>{\columncolor{shadegray}}c}{
\rotatebox{90}{\strut \textit{Phaet}}} & \multicolumn{1}{c}{\rotatebox{90}{\strut GenBio-PathFM}} & \multicolumn{1}{>{\columncolor{shadegray}}c}{
\rotatebox{90}{\strut GenBio-PathFM$^\dagger$}} & \multicolumn{1}{c}{\rotatebox{90}{\strut Phikon}} & \multicolumn{1}{>{\columncolor{shadegray}}c}{
\rotatebox{90}{\strut Phikon$^\dagger$}} \\
\midrule
\endfirsthead
\toprule
\textbf{[Dataset][Task][Metric]} & \multicolumn{1}{c}{\rotatebox{90}{\strut Midnight-12k}} & \multicolumn{1}{>{\columncolor{shadegray}}c}{
\rotatebox{90}{\strut Midnight-12k$^\dagger$}} & \multicolumn{1}{c}{\rotatebox{90}{\strut UNI2-h}} & \multicolumn{1}{>{\columncolor{shadegray}}c}{
\rotatebox{90}{\strut UNI2-h$^\dagger$}} & \multicolumn{1}{c}{\rotatebox{90}{\strut AquaViT}} & \multicolumn{1}{>{\columncolor{shadegray}}c}{
\rotatebox{90}{\strut AquaViT$^\dagger$}} & \multicolumn{1}{c}{\rotatebox{90}{\strut H0-mini}} & \multicolumn{1}{>{\columncolor{shadegray}}c}{
\rotatebox{90}{\strut H0-mini$^\dagger$}} & \multicolumn{1}{c}{\rotatebox{90}{\strut H-Optimus-0}} & \multicolumn{1}{>{\columncolor{shadegray}}c}{
\rotatebox{90}{\strut H-Optimus-0$^\dagger$}} & \multicolumn{1}{c}{\rotatebox{90}{\strut Virchow2}} & \multicolumn{1}{>{\columncolor{shadegray}}c}{
\rotatebox{90}{\strut Virchow2$^\dagger$}} & \multicolumn{1}{c}{\rotatebox{90}{\strut Prov-GigaPath}} & \multicolumn{1}{>{\columncolor{shadegray}}c}{
\rotatebox{90}{\strut Prov-GigaPath$^\dagger$}} & \multicolumn{1}{c}{\rotatebox{90}{\strut Phikon-v2}} & \multicolumn{1}{>{\columncolor{shadegray}}c}{
\rotatebox{90}{\strut Phikon-v2$^\dagger$}} & \multicolumn{1}{c}{\rotatebox{90}{\strut GenBio-PathFM}} & \multicolumn{1}{>{\columncolor{shadegray}}c}{
\rotatebox{90}{\strut GenBio-PathFM$^\dagger$}} & \multicolumn{1}{c}{\rotatebox{90}{\strut Phikon}} & \multicolumn{1}{>{\columncolor{shadegray}}c}{
\rotatebox{90}{\strut Phikon$^\dagger$}} \\
\midrule
\endhead
\midrule \multicolumn{21}{r}{\textit{continued\ldots}} \\
\endfoot
\bottomrule
\endlastfoot
\multicolumn{21}{l}{\textbf{Molecular}} \\
\cmidrule{1-21}
{[BC Therapy][HER2 status][AUC]} & 58.9 & 59.6 & 66.6 & 63.8 & 65.8 & 64.7 & 66.8 & 65.7 & 58.6 & 63.6 & 65.2 & 64.8 & 62.7 & 65.3 & \textbf{69.5} & 66.6 & 65.6 & 66.4 & 62.8 & 66.0 \\
{[CPTAC BRCA][PIK3CA][AUC]} & 61.3 & 60.7 & 57.6 & 65.4 & 63.2 & \textbf{66.4} & 62.9 & 59.8 & 62.7 & 62.7 & 58.4 & 65.3 & 60.9 & 61.3 & 57.3 & 56.6 & 60.9 & 62.8 & 58.7 & 58.0 \\
{[CPTAC BRCA][TP53][AUC]} & 77.4 & 83.4 & 77.8 & 77.2 & 78.8 & 79.7 & 79.2 & 79.9 & 75.1 & 80.8 & 78.7 & 78.4 & 77.1 & 78.8 & \textbf{85.4} & 85.2 & 78.6 & 79.7 & 82.4 & 81.8 \\
{[CPTAC CCRCC][BAP1][AUC]} & 67.6 & 71.4 & 63.9 & 68.7 & 64.6 & 68.6 & 66.7 & 71.7 & 61.5 & 69.6 & 64.6 & 70.7 & 66.1 & \textbf{76.0} & 62.3 & 66.1 & 66.0 & 69.9 & 61.5 & 65.9 \\
{[CPTAC CCRCC][PBRM1][AUC]} & 43.2 & 55.1 & 40.5 & 44.5 & 50.0 & 52.8 & 53.5 & 52.2 & 44.6 & 46.6 & 50.6 & 53.9 & 54.0 & 53.6 & \textbf{57.7} & 56.9 & 51.9 & 54.5 & 47.8 & 47.9 \\
{[CPTAC CCRCC][VHL][AUC]} & 52.5 & 49.7 & 51.7 & 50.4 & 52.3 & 50.2 & 52.5 & 52.1 & 47.6 & 51.1 & 52.4 & 50.3 & \textbf{54.0} & 53.5 & 42.6 & 48.9 & 51.0 & 53.2 & 45.2 & 45.6 \\
{[CPTAC COAD][ACVR2A][AUC]} & 72.9 & 78.8 & 81.5 & 80.1 & 81.7 & 84.0 & 82.8 & 82.9 & 83.1 & \textbf{85.2} & 77.5 & 79.0 & 78.3 & 83.0 & 77.2 & 80.9 & 77.0 & 79.4 & 73.7 & 77.1 \\
{[CPTAC COAD][APC][AUC]} & 67.2 & 70.2 & 74.5 & \textbf{76.8} & 75.6 & 76.7 & 73.9 & 76.4 & 70.9 & 75.0 & 71.8 & 72.5 & 72.4 & 75.0 & 72.7 & 70.7 & 73.4 & 74.6 & 70.6 & 72.4 \\
{[CPTAC COAD][ARID1A][AUC]} & 72.0 & 70.3 & 68.1 & 71.2 & 73.5 & 73.2 & 72.8 & 74.0 & 77.9 & 73.6 & 65.2 & 74.5 & 70.7 & 68.5 & 74.0 & \textbf{78.0} & 72.6 & 74.9 & 69.7 & 71.7 \\
{[CPTAC COAD][KRAS][AUC]} & 58.5 & 61.0 & 69.9 & 66.9 & 64.5 & 61.8 & 67.7 & 67.2 & 59.1 & 67.2 & 65.0 & 65.8 & 62.8 & 68.5 & 70.8 & \textbf{70.8} & 59.5 & 62.6 & 59.6 & 64.1 \\
{[CPTAC COAD][MSI-H][AUC]} & 81.2 & 83.7 & 89.9 & 90.4 & 88.8 & 91.0 & 89.3 & 92.2 & 86.5 & 92.6 & 84.4 & 89.8 & 85.8 & \textbf{92.7} & 82.9 & 85.2 & 85.1 & 88.4 & 81.2 & 84.9 \\
{[CPTAC COAD][PIK3CA][AUC]} & 58.1 & 58.3 & 55.1 & 60.1 & 63.8 & 64.9 & 63.2 & 63.8 & 67.4 & \textbf{69.9} & 64.5 & 65.2 & 61.0 & 66.5 & 58.3 & 58.2 & 61.2 & 63.2 & 60.0 & 61.1 \\
{[CPTAC COAD][SETD1B][AUC]} & 80.0 & 76.9 & 84.5 & 85.1 & 87.7 & 85.4 & 86.7 & 84.4 & 84.4 & \textbf{88.7} & 78.7 & 80.2 & 80.2 & 86.3 & 75.4 & 77.1 & 78.9 & 79.7 & 76.9 & 80.5 \\
{[CPTAC COAD][TP53][AUC]} & 64.9 & 67.8 & 69.4 & 69.2 & 68.7 & 66.6 & 68.0 & 67.3 & 67.1 & 70.2 & 65.0 & 66.5 & 66.4 & 72.8 & \textbf{73.6} & 69.4 & 57.7 & 60.9 & 69.5 & 69.9 \\
{[CPTAC GBM][EGFR][AUC]} & 52.7 & 61.8 & 57.7 & 62.3 & 59.7 & 64.1 & 57.1 & 61.5 & 53.8 & 63.2 & 61.5 & 65.7 & 60.1 & \textbf{66.8} & 52.9 & 56.8 & 62.6 & 66.6 & 52.9 & 61.7 \\
{[CPTAC GBM][TP53][AUC]} & 73.9 & 72.9 & 81.2 & 84.6 & \textbf{87.6} & 83.9 & 84.3 & 77.6 & 80.4 & 82.2 & 70.9 & 73.0 & 84.8 & 84.7 & 75.6 & 77.9 & 75.4 & 77.2 & 75.3 & 84.6 \\
{[CPTAC HNSC][CASP8][AUC]} & 60.8 & \textbf{76.1} & 65.2 & 63.8 & 63.4 & 65.7 & 65.8 & 68.3 & 60.4 & 71.0 & 57.7 & 63.8 & 63.2 & 68.7 & 54.6 & 48.5 & 55.4 & 59.8 & 55.8 & 59.6 \\
{[CPTAC LSCC][ARID1A][AUC]} & 50.3 & \textbf{61.5} & 44.8 & 51.5 & 49.9 & 54.1 & 47.7 & 49.9 & 46.0 & 53.0 & 47.2 & 52.9 & 45.4 & 51.2 & 39.7 & 49.8 & 52.3 & 50.4 & 43.7 & 47.3 \\
{[CPTAC LSCC][KEAP1][AUC]} & 60.5 & 61.7 & 63.8 & 60.6 & 63.9 & 64.2 & 62.1 & 61.4 & 63.6 & 65.3 & 62.8 & 62.4 & 57.7 & 64.5 & 64.6 & \textbf{67.1} & 59.9 & 61.7 & 60.3 & 64.5 \\
{[CPTAC LUAD][EGFR][AUC]} & 77.5 & 77.4 & 80.4 & 82.2 & 84.5 & 84.3 & \textbf{85.2} & 83.1 & 83.2 & 82.9 & 80.9 & 76.1 & 82.7 & 81.9 & 75.1 & 80.2 & 76.5 & 76.6 & 77.6 & 82.0 \\
{[CPTAC LUAD][KRAS][AUC]} & 65.7 & 69.1 & 67.4 & 68.5 & 73.9 & 73.7 & 66.6 & 71.5 & 72.1 & 73.1 & \textbf{74.2} & 68.8 & 70.9 & 72.0 & 65.9 & 70.6 & 64.9 & 62.9 & 68.9 & 68.2 \\
{[CPTAC LUAD][STK11][AUC]} & 88.0 & 86.7 & 89.6 & \textbf{90.5} & 88.7 & 89.5 & 88.1 & 87.9 & 89.4 & 89.4 & 82.8 & 83.0 & 87.8 & 88.2 & 86.5 & 88.1 & 86.4 & 86.7 & 87.2 & 88.1 \\
{[CPTAC LUAD][TP53][AUC]} & 70.1 & 67.9 & \textbf{74.3} & 73.0 & 70.2 & 71.1 & 71.8 & 69.8 & 68.4 & 69.6 & 74.1 & 70.6 & 72.1 & 71.6 & 67.3 & 70.3 & 68.6 & 69.9 & 71.7 & 73.1 \\
{[CPTAC PDA][SMAD4][AUC]} & 39.7 & 38.5 & 32.3 & 37.8 & 39.8 & 40.2 & 41.3 & 40.5 & 41.0 & 42.5 & \textbf{43.5} & 40.7 & 37.3 & 41.7 & 37.4 & 39.9 & 39.2 & 39.9 & 37.0 & 39.7 \\
{[CPTAC UCEC][CTNNB1][AUC]} & 75.2 & 73.7 & \textbf{79.5} & 77.8 & 75.7 & 74.8 & 76.6 & 75.2 & 76.5 & 76.7 & 73.7 & 71.0 & 74.3 & 73.1 & 78.9 & 78.0 & 71.7 & 72.0 & 73.3 & 72.0 \\
{[CPTAC UCEC][PTEN][AUC]} & 53.8 & 64.0 & 66.2 & 66.6 & 67.6 & 66.8 & 65.9 & 63.0 & 66.4 & 67.2 & 66.5 & 60.2 & 61.2 & 67.2 & \textbf{72.0} & 65.3 & 58.7 & 63.0 & 63.7 & 64.6 \\
{[Mut-Het-RCC][BAP1][AUC]} & 86.5 & 88.4 & 89.1 & 89.9 & 89.9 & \textbf{91.0} & 90.1 & 89.8 & 88.8 & 89.6 & 86.6 & 87.8 & 87.3 & 88.4 & 88.5 & 88.4 & 87.0 & 88.0 & 88.3 & 88.5 \\
{[Mut-Het-RCC][PBRM1][AUC]} & 80.8 & 82.0 & 81.7 & 82.8 & 82.4 & 82.6 & 80.9 & 80.6 & 79.4 & 81.0 & 81.8 & 81.8 & 78.2 & 81.1 & 81.5 & \textbf{83.0} & 82.8 & 82.9 & 79.4 & 80.5 \\
{[Mut-Het-RCC][SETD2][AUC]} & 71.0 & \textbf{74.9} & 70.1 & 70.4 & 70.8 & 71.2 & 69.7 & 70.1 & 66.3 & 68.0 & 69.9 & 70.7 & 63.3 & 65.4 & 72.1 & 72.9 & 71.2 & 72.6 & 69.6 & 73.2 \\
\cmidrule{1-21}
\textit{Average} & 66.3 & \cellcolor{green!35}69.1 & 68.8 & \cellcolor{green!35}70.1 & 70.6 & \cellcolor{green!35}71.1 & 70.3 & \cellcolor{green!35}70.3 & 68.4 & \cellcolor{green!35}\textbf{71.4} & 68.1 & \cellcolor{green!35}69.1 & 68.2 & \cellcolor{green!35}71.3 & 68.0 & \cellcolor{green!35}69.2 & 67.3 & \cellcolor{green!35}69.0 & 66.4 & \cellcolor{green!35}68.8 \\
\addlinespace[4pt]
\midrule
\multicolumn{21}{l}{\textbf{Morphology}} \\
\cmidrule{1-21}
{[BC Therapy][Grading][$\kappa$]} & 27.0 & \textbf{38.0} & 34.4 & 32.2 & 33.1 & 31.5 & 35.2 & 37.3 & 27.5 & 27.9 & 33.1 & 32.6 & 29.1 & 36.1 & 33.5 & 35.4 & 30.5 & 33.2 & 28.7 & 35.1 \\
{[IMP Cervical][Morph. subtyping][bAcc]} & 81.1 & 79.0 & 80.1 & 80.1 & \textbf{81.3} & 79.9 & 81.3 & 80.5 & 80.9 & 80.4 & 79.7 & 78.9 & 78.1 & 80.6 & 70.2 & 76.4 & 77.3 & 74.2 & 77.0 & 78.0 \\
{[CPTAC BRCA][Immune class][bAcc]} & 52.5 & 51.0 & 55.1 & 51.2 & 53.0 & 51.4 & 52.3 & 52.1 & 50.8 & 49.4 & 52.9 & \textbf{55.2} & 52.5 & 54.2 & 52.5 & 54.8 & 49.7 & 50.2 & 50.5 & 52.1 \\
{[CPTAC CCRCC][Immune class][bAcc]} & 36.0 & 37.4 & 32.8 & 34.9 & 38.5 & 38.7 & 38.6 & \textbf{39.8} & 33.6 & 34.3 & 34.3 & 34.3 & 35.6 & 35.8 & 33.0 & 34.4 & 35.5 & 38.6 & 37.1 & 38.5 \\
{[CPTAC COAD][Immune class][bAcc]} & 38.7 & \textbf{44.7} & 40.3 & 43.3 & 34.2 & 37.2 & 37.7 & 42.0 & 34.3 & 35.6 & 44.2 & 43.5 & 38.1 & 42.2 & 38.0 & 43.5 & 38.1 & 38.7 & 35.3 & 35.4 \\
{[CPTAC GBM][Immune class][bAcc]} & 54.1 & \textbf{55.1} & 51.9 & 53.3 & 48.9 & 50.4 & 50.2 & 53.0 & 50.8 & 54.1 & 46.0 & 51.0 & 51.5 & 54.8 & 47.5 & 49.3 & 48.8 & 54.2 & 44.6 & 47.4 \\
{[CPTAC HNSC][Histologic grade][$\kappa$]} & 30.1 & \textbf{35.0} & 23.8 & 27.5 & 25.1 & 27.1 & 25.4 & 23.6 & 23.1 & 25.0 & 29.2 & 26.6 & 26.9 & 25.4 & 26.8 & 15.0 & 19.2 & 28.0 & 30.7 & 25.4 \\
{[CPTAC HNSC][Immune class][bAcc]} & 40.6 & 46.2 & 44.5 & 47.8 & 49.5 & 51.1 & 46.2 & 48.4 & 43.8 & 48.8 & 48.7 & \textbf{52.2} & 44.0 & 51.2 & 42.2 & 42.4 & 43.9 & 49.0 & 41.4 & 45.8 \\
{[CPTAC LSCC][Histologic grade][$\kappa$]} & 14.1 & 10.8 & 25.9 & 25.1 & 22.0 & 20.9 & 18.5 & 16.2 & \textbf{28.0} & 26.2 & 22.2 & 18.6 & 22.4 & 16.5 & 16.2 & 16.0 & 14.0 & 15.4 & 21.8 & 19.0 \\
{[CPTAC LSCC][Immune class][bAcc]} & 57.6 & 58.8 & 49.8 & 53.7 & 51.0 & 52.3 & 58.3 & 57.9 & 52.8 & 56.6 & 58.9 & \textbf{59.8} & 53.7 & 55.9 & 54.0 & 55.8 & 56.5 & 59.0 & 49.6 & 54.0 \\
{[CPTAC LUAD][Immune class][bAcc]} & 48.6 & 54.8 & 47.0 & 48.4 & 49.0 & 49.7 & 49.6 & 50.1 & 47.1 & 49.7 & 51.9 & 54.0 & 46.8 & 48.9 & 46.0 & 46.4 & 55.0 & \textbf{59.1} & 44.7 & 49.5 \\
{[CPTAC PDA][Immune class][bAcc]} & \textbf{43.8} & 43.3 & 40.2 & 40.4 & 39.4 & 40.0 & 39.0 & 39.1 & 39.0 & 40.1 & 41.5 & 40.8 & 36.5 & 41.9 & 37.1 & 38.4 & 41.8 & 41.1 & 38.3 & 42.5 \\
{[CPTAC UCEC][Immune class][bAcc]} & 39.8 & 41.1 & 38.6 & 40.2 & 37.1 & 42.7 & 41.3 & \textbf{47.6} & 38.0 & 40.2 & 42.2 & 42.7 & 32.3 & 42.8 & 34.9 & 37.1 & 36.6 & 42.4 & 32.7 & 37.7 \\
{[Hancock][Grading: SCC Keratinizing][$\kappa$]} & 26.4 & 32.6 & 31.2 & 34.4 & 31.6 & 32.4 & 36.2 & 32.6 & \textbf{38.1} & 34.2 & 31.2 & 31.4 & 33.5 & 33.1 & 34.7 & 36.6 & 27.5 & 28.0 & 29.3 & 32.8 \\
{[Hancock][Grading: SCC Non-Kerat.][$\kappa$]} & 22.9 & 23.2 & 23.9 & \textbf{29.1} & 12.9 & 24.5 & 19.4 & 20.7 & 17.5 & 23.1 & 18.8 & 27.3 & 18.3 & 26.4 & 9.4 & 20.9 & 19.0 & 22.0 & 5.8 & 12.9 \\
{[Hancock][Lymph. invasion][AUC]} & 66.9 & 69.1 & 63.8 & 67.4 & 67.5 & 68.3 & 68.6 & 67.0 & 66.2 & 68.8 & 65.3 & 67.1 & 64.0 & 67.5 & 61.9 & 64.6 & 67.7 & \textbf{69.7} & 61.5 & 65.2 \\
{[Hancock][Perineural invasion][AUC]} & 73.5 & \textbf{73.8} & 68.5 & 68.6 & 70.9 & 69.2 & 69.8 & 70.2 & 73.2 & 71.7 & 72.4 & 71.7 & 70.5 & 71.7 & 66.9 & 70.3 & 72.8 & 72.8 & 68.1 & 67.9 \\
{[Hancock][Primary vs. Metastasis][AUC]} & 64.0 & 70.9 & 67.0 & 67.0 & 66.7 & 66.8 & 64.5 & 66.4 & 64.9 & 64.1 & 64.5 & 64.7 & 64.3 & 65.7 & 67.4 & \textbf{71.8} & 68.8 & 67.5 & 66.7 & 68.9 \\
{[Hancock][Primary tumor site][bAcc]} & 73.8 & \textbf{74.2} & 73.9 & 74.2 & 73.4 & 72.6 & 72.4 & 70.6 & 73.6 & 73.8 & 73.6 & 72.1 & 72.3 & 70.9 & 74.0 & 71.9 & 73.2 & 73.6 & 74.1 & 72.6 \\
{[Hancock][Vascular invasion][AUC]} & 61.1 & 63.0 & 67.5 & 70.4 & 71.6 & 68.8 & 66.9 & 65.4 & 68.1 & 64.7 & 70.3 & \textbf{71.6} & 66.3 & 65.6 & 58.6 & 61.9 & 67.6 & 67.9 & 63.5 & 61.8 \\
{[Post-NAT BRCA][Lymph. invasion][AUC]} & 51.2 & 49.0 & 55.8 & \textbf{60.6} & 59.2 & 59.8 & 56.5 & 58.6 & 59.7 & 56.8 & 58.7 & 57.0 & 52.3 & 59.6 & 59.5 & 57.9 & 58.1 & 55.1 & 46.4 & 56.7 \\
{[PANDA][Grading: ISUP grade][$\kappa$]} & 95.4 & 95.1 & 96.5 & 96.6 & \textbf{97.2} & 96.9 & 96.4 & 96.5 & 96.7 & 96.4 & 95.7 & 95.9 & 96.2 & 97.0 & 95.9 & 95.7 & 95.6 & 95.8 & 96.5 & 95.7 \\
\cmidrule{1-21}
\textit{Average} & 50.0 & \cellcolor{green!35}52.1 & 50.6 & \cellcolor{green!35}52.1 & 50.6 & \cellcolor{green!35}51.5 & 51.1 & \cellcolor{green!35}51.6 & 50.3 & \cellcolor{green!35}51.0 & 51.6 & \cellcolor{green!35}\textbf{52.2} & 49.3 & \cellcolor{green!35}52.0 & 48.2 & \cellcolor{green!35}49.9 & 49.9 & \cellcolor{green!35}51.6 & 47.5 & \cellcolor{green!35}49.8 \\
\addlinespace[4pt]
\midrule
\multicolumn{21}{l}{\textbf{Survival}} \\
\cmidrule{1-21}
{[Boehmk][Survival: PFS][C-idx]} & 53.3 & \textbf{54.7} & 50.8 & 54.2 & 51.6 & 52.1 & 51.3 & 52.1 & 49.8 & 51.0 & 49.5 & 53.3 & 53.7 & 53.6 & 49.5 & 52.8 & 48.6 & 48.9 & 50.3 & 49.8 \\
{[CPTAC CCRCC][Survival: OS][C-idx]} & 67.0 & 68.3 & 66.3 & 65.3 & 59.8 & 59.6 & 60.1 & 60.3 & 57.4 & 55.6 & 60.0 & 64.4 & 57.7 & 62.4 & 62.8 & \textbf{69.2} & 54.7 & 54.3 & 63.1 & 61.2 \\
{[CPTAC HNSC][Survival: OS][C-idx]} & 62.1 & 61.3 & 55.7 & 60.7 & 57.8 & 55.4 & 57.8 & 59.2 & 61.2 & 58.6 & 53.5 & 58.0 & 53.9 & 56.8 & \textbf{68.3} & 67.0 & 55.7 & 59.5 & 65.2 & 62.5 \\
{[CPTAC LUAD][Survival: OS][C-idx]} & 53.1 & 57.8 & 52.2 & 52.0 & 55.2 & \textbf{63.6} & 59.0 & 53.3 & 46.9 & 51.5 & 56.6 & 51.6 & 47.3 & 45.8 & 55.2 & 54.7 & 51.6 & 50.9 & 54.9 & 49.2 \\
{[CPTAC PDA][Survival: OS][C-idx]} & 49.5 & \textbf{55.0} & 48.7 & 50.8 & 50.7 & 50.1 & 48.3 & 47.3 & 51.3 & 50.5 & 48.2 & 50.7 & 47.9 & 51.2 & 52.3 & 47.0 & 49.5 & 51.5 & 50.9 & 48.0 \\
{[Hancock][Survival: OS Ttt. Rdc][C-idx]} & 59.5 & 58.1 & 58.6 & \textbf{60.6} & 54.9 & 58.1 & 58.2 & 58.7 & 55.6 & 56.9 & 58.8 & 54.5 & 54.4 & 56.0 & 57.1 & 57.9 & 50.2 & 50.6 & 53.6 & 57.1 \\
{[MBC][Survival: OS][C-idx]} & 53.7 & 54.2 & 50.7 & 50.0 & 50.8 & 46.0 & 48.6 & 51.5 & 55.3 & \textbf{57.4} & 54.0 & 55.1 & 49.3 & 48.5 & 45.6 & 50.3 & 48.8 & 49.7 & 47.3 & 47.6 \\
\cmidrule{1-21}
\textit{Average} & 56.9 & \cellcolor{green!35}\textbf{58.5} & 54.7 & \cellcolor{green!35}56.2 & 54.4 & \cellcolor{green!35}55.0 & 54.8 & \cellcolor{red!35}54.6 & 53.9 & \cellcolor{green!35}54.5 & 54.4 & \cellcolor{green!35}55.4 & 52.0 & \cellcolor{green!35}53.5 & 55.8 & \cellcolor{green!35}57.0 & 51.3 & \cellcolor{green!35}52.2 & 55.0 & \cellcolor{red!35}53.6 \\
\addlinespace[4pt]
\midrule
\multicolumn{21}{l}{\textbf{Treatment response}} \\
\cmidrule{1-21}
{[BC Therapy][ER status][AUC]} & 51.5 & 68.5 & 67.9 & 69.7 & 70.7 & 69.9 & 71.4 & 73.2 & 59.0 & 71.8 & 67.5 & 68.5 & 73.1 & 71.3 & 69.8 & 68.7 & 71.9 & \textbf{73.3} & 63.9 & 68.3 \\
{[BC Therapy][Res. cancer burden][bAcc]} & 26.2 & 29.6 & 29.4 & 29.0 & 29.6 & 29.1 & 29.7 & 29.0 & 27.3 & \textbf{30.5} & 29.2 & 28.3 & 29.6 & 28.4 & 28.7 & 29.8 & 26.8 & 28.9 & 28.7 & 30.0 \\
{[MBC][RECIST][$\kappa$]} & 15.4 & 14.8 & 7.8 & 11.6 & 12.8 & 17.5 & \textbf{19.6} & 15.7 & 13.6 & 16.0 & 14.5 & 15.6 & 18.1 & 16.8 & 12.3 & 17.1 & 8.5 & 14.4 & 12.8 & 13.4 \\
{[NADT Prostate][Ttt. response][AUC]} & 73.1 & 84.1 & 79.8 & 84.2 & 80.4 & 84.1 & 83.7 & \textbf{85.1} & 75.5 & 80.2 & 74.9 & 79.1 & 65.7 & 81.3 & 72.7 & 71.9 & 76.0 & 77.1 & 65.8 & 75.4 \\
{[OV Bevacizumab][Ttt. response][AUC]} & 60.3 & \textbf{63.8} & 56.6 & 60.2 & 52.2 & 54.1 & 48.6 & 52.7 & 48.8 & 53.5 & 51.6 & 60.5 & 54.7 & 51.9 & 37.9 & 48.9 & 54.9 & 58.4 & 49.9 & 52.0 \\
\cmidrule{1-21}
\textit{Average} & 45.3 & \cellcolor{green!35}\textbf{52.2} & 48.3 & \cellcolor{green!35}50.9 & 49.2 & \cellcolor{green!35}50.9 & 50.6 & \cellcolor{green!35}51.1 & 44.8 & \cellcolor{green!35}50.4 & 47.5 & \cellcolor{green!35}50.4 & 48.2 & \cellcolor{green!35}49.9 & 44.3 & \cellcolor{green!35}47.3 & 47.6 & \cellcolor{green!35}50.4 & 44.2 & \cellcolor{green!35}47.8 \\
\addlinespace[4pt]
\midrule
\textit{Grand Average} & 54.6 & \cellcolor{green!35}\textbf{58.0} & 55.6 & \cellcolor{green!35}57.3 & 56.2 & \cellcolor{green!35}57.1 & 56.7 & \cellcolor{green!35}56.9 & 54.4 & \cellcolor{green!35}56.8 & 55.4 & \cellcolor{green!35}56.8 & 54.5 & \cellcolor{green!35}56.7 & 54.1 & \cellcolor{green!35}55.8 & 54.0 & \cellcolor{green!35}55.8 & 53.3 & \cellcolor{green!35}55.0 \\
\end{longtable}
}

%% file: tables/inc/thunder_full_redacted.tex
{\footnotesize\setlength{\tabcolsep}{3pt}
\begin{longtable}{@{}lrrrrrrr@{}}
\caption{THUNDER benchmark: extended leaderboard. Per-task scores with unified rank in parentheses; rank sum~($\downarrow$) is the primary metric. Shaded rows (\textbf{model}$^\dagger$) report fine-tuned results. $^{\mathrm{MP}}$~denotes a base model evaluated with mixed precision (our setup); for each base model, only the better-performing precision variant is shown. Unlabelled rows report full-precision results from the official leaderboard. Sorted by rank sum (ascending).} \label{tab:thunder_full} \\
\toprule
\textbf{Model} & \multicolumn{1}{c}{{\strut KNN~$\uparrow$}} & \multicolumn{1}{c}{{\strut Lin.~prob.~$\uparrow$}} & \multicolumn{1}{c}{{\strut Few-shot~$\uparrow$}} & \multicolumn{1}{c}{{\strut Seg.~$\uparrow$}} & \multicolumn{1}{c}{{\strut Calib.~$\downarrow$}} & \multicolumn{1}{c}{{\strut Adv.~att.~$\downarrow$}} & \multicolumn{1}{c}{{\strut Rank~sum~$\downarrow$}} \\
\midrule
\endfirsthead
\toprule
\textbf{Model} & \multicolumn{1}{c}{{\strut KNN~$\uparrow$}} & \multicolumn{1}{c}{{\strut Lin.~prob.~$\uparrow$}} & \multicolumn{1}{c}{{\strut Few-shot~$\uparrow$}} & \multicolumn{1}{c}{{\strut Seg.~$\uparrow$}} & \multicolumn{1}{c}{{\strut Calib.~$\downarrow$}} & \multicolumn{1}{c}{{\strut Adv.~att.~$\downarrow$}} & \multicolumn{1}{c}{{\strut Rank~sum~$\downarrow$}} \\
\midrule
\endhead
\midrule \multicolumn{8}{r}{\textit{continued\ldots}} \\
\endfoot
\bottomrule
\endlastfoot
\rowcolor{shadegray}UNI2-h$^\dagger$ & 83.4~(3) & 85.5~(2) & 79.5~(3) & 67.6~(18) & 2.5~(2) & 24.1~(3) & \textbf{31} \\
UNI2-h$^{\mathrm{MP}}$ & 83.3~(4) & \textbf{86.3~(1)} & \textbf{79.8~(1)} & 68.1~(10) & 3.7~(10) & 31.0~(9) & 35 \\
\rowcolor{shadegray}H-Optimus-0$^\dagger$ & 81.9~(7) & 84.0~(10) & 77.4~(6) & 68.1~(10) & 3.2~(5) & 32.4~(12) & 50 \\
\rowcolor{shadegray}GenBio-PathFM$^\dagger$ & \textbf{83.9~(1)} & 85.3~(3) & 79.6~(2) & 66.8~(23) & 4.0~(17) & 26.2~(5) & 51 \\
\rowcolor{shadegray}\textit{Mascaret} & 81.7~(8) & 84.6~(9) & 75.2~(15) & 67.6~(18) & \textbf{2.3~(1)} & 23.2~(2) & 53 \\
\rowcolor{shadegray}Virchow2$^\dagger$ & 82.6~(6) & 85.1~(4) & 76.6~(9) & 68.0~(12) & 4.2~(21) & \textbf{7.7~(1)} & 53 \\
GenBio-PathFM$^{\mathrm{MP}}$ & 83.5~(2) & 85.1~(4) & 79.4~(4) & 67.2~(22) & 3.7~(10) & 32.7~(13) & 55 \\
Virchow2 & 82.9~(5) & 84.8~(6) & 73.9~(20) & \textbf{69.3~(1)} & 3.9~(15) & 31.1~(10) & 57 \\
\rowcolor{shadegray}Prov-GigaPath$^\dagger$ & 80.8~(11) & 83.0~(16) & 76.7~(8) & 65.8~(27) & 3.2~(5) & 28.6~(7) & 74 \\
Midnight-12k & 79.9~(15) & 84.7~(7) & 71.5~(30) & 68.8~(5) & 2.9~(3) & 37.0~(16) & 76 \\
H0-mini & 79.7~(17) & 83.8~(11) & 75.0~(18) & 69.1~(3) & 3.8~(13) & 34.3~(15) & 77 \\
UNI & 80.8~(11) & 83.5~(14) & 78.1~(5) & 67.8~(16) & 3.8~(13) & 40.3~(22) & 81 \\
KEEP & 81.5~(9) & 83.2~(15) & 77.1~(7) & 68.0~(12) & 4.0~(17) & 44.9~(27) & 87 \\
\rowcolor{shadegray}AquaViT$^\dagger$ & 80.1~(14) & 82.4~(20) & 75.1~(16) & 68.3~(8) & 5.3~(28) & 29.3~(8) & 94 \\
AquaViT$^{\mathrm{MP}}$ & 80.5~(13) & 82.7~(19) & 75.7~(13) & 68.0~(12) & 5.2~(27) & 32.0~(11) & 95 \\
H-Optimus-0$^{\mathrm{MP}}$ & 81.5~(9) & 83.7~(12) & 76.2~(12) & 63.5~(30) & 3.6~(9) & 42.1~(23) & 95 \\
\rowcolor{shadegray}H0-mini$^\dagger$ & 78.7~(22) & 82.4~(20) & 73.8~(21) & 67.5~(20) & 3.7~(10) & 27.6~(6) & 99 \\
Hibou-B & 78.9~(20) & 81.2~(26) & 76.3~(11) & 67.8~(16) & 3.2~(5) & 52.7~(28) & 106 \\
OpenMidnight & 79.3~(19) & 84.7~(7) & 43.7~(31) & 69.1~(3) & 5.4~(29) & 38.3~(17) & 106 \\
Hibou-L & 78.6~(24) & 83.7~(12) & 73.8~(21) & 68.6~(7) & 4.7~(25) & 39.5~(21) & 110 \\
Prov-GigaPath & 79.5~(18) & 82.9~(17) & 75.5~(14) & 63.5~(30) & 3.4~(8) & 42.1~(23) & 110 \\
CONCH~1.5 & 79.9~(15) & 82.4~(20) & 75.0~(18) & 68.8~(5) & 4.6~(23) & 75.8~(31) & 112 \\
Virchow & 77.4~(28) & 82.8~(18) & 71.8~(28) & 69.2~(2) & 4.5~(22) & 38.3~(17) & 115 \\
CONCH & 78.8~(21) & 81.9~(23) & 73.4~(24) & 68.3~(8) & 4.1~(20) & 57.3~(29) & 125 \\
Kaiko ViT-B/16 & 78.7~(22) & 81.4~(25) & 76.4~(10) & 66.8~(23) & 5.0~(26) & 38.8~(19) & 125 \\
\rowcolor{shadegray}\textit{Phaet} & 77.7~(26) & 80.7~(29) & 73.3~(25) & 65.3~(28) & 3.0~(4) & 38.8~(19) & 131 \\
Kaiko ViT-S/16 & 78.2~(25) & 81.7~(24) & 75.1~(16) & 66.8~(23) & 4.6~(23) & 42.5~(25) & 136 \\
Phikon & 75.7~(30) & 80.9~(28) & 73.6~(23) & 68.0~(12) & 5.8~(31) & 33.5~(14) & 138 \\
\rowcolor{shadegray}Phikon$^\dagger$ & 76.1~(29) & 79.8~(30) & 73.2~(26) & 66.6~(26) & 5.4~(29) & 24.5~(4) & 144 \\
Phikon-v2 & 73.9~(31) & 79.7~(31) & 71.8~(28) & 67.4~(21) & 3.9~(15) & 43.8~(26) & 152 \\
MUSK & 77.7~(26) & 81.1~(27) & 71.9~(27) & 65.1~(29) & 4.0~(17) & 71.9~(30) & 156 \\
\end{longtable}
}

%% file: tables/inc/hest_cls_full_redacted.tex
{\scriptsize\setlength{\tabcolsep}{3pt}
\begin{longtable}{@{}lrrrrrrrrrr@{}}
\caption{HEST benchmark: extended leaderboard (Pearson correlation~$\uparrow$). Shaded rows (\textbf{model}$^\dagger$) report fine-tuned results; unshaded rows report base model results from the HEST leaderboard. Sorted by average (descending).} \label{tab:hest_cls_full} \\
\toprule
\textbf{Model} & \multicolumn{1}{c}{{\strut IDC}} & \multicolumn{1}{c}{{\strut PRAD}} & \multicolumn{1}{c}{{\strut PAAD}} & \multicolumn{1}{c}{{\strut SKCM}} & \multicolumn{1}{c}{{\strut COAD}} & \multicolumn{1}{c}{{\strut READ}} & \multicolumn{1}{c}{{\strut CCRCC}} & \multicolumn{1}{c}{{\strut LUNG}} & \multicolumn{1}{c}{{\strut LYMPH$_{\text{IDC}}$}} & \textbf{Average} \\
\midrule
\endfirsthead
\toprule
\textbf{Model} & \multicolumn{1}{c}{{\strut IDC}} & \multicolumn{1}{c}{{\strut PRAD}} & \multicolumn{1}{c}{{\strut PAAD}} & \multicolumn{1}{c}{{\strut SKCM}} & \multicolumn{1}{c}{{\strut COAD}} & \multicolumn{1}{c}{{\strut READ}} & \multicolumn{1}{c}{{\strut CCRCC}} & \multicolumn{1}{c}{{\strut LUNG}} & \multicolumn{1}{c}{{\strut LYMPH$_{\text{IDC}}$}} & \textbf{Average} \\
\midrule
\endhead
\midrule \multicolumn{11}{r}{\textit{continued\ldots}} \\
\endfoot
\bottomrule
\endlastfoot
\rowcolor{shadegray}H-Optimus-0$^\dagger$ & \textbf{0.6069} & 0.3695 & 0.5205 & 0.6764 & 0.3136 & \textbf{0.2466} & \textbf{0.2791} & \textbf{0.5811} & 0.2671 & \textbf{0.4290} \\
\rowcolor{shadegray}UNI2-h$^\dagger$ & 0.6016 & 0.3763 & \textbf{0.5290} & \textbf{0.6781} & 0.3294 & 0.2239 & 0.2734 & 0.5729 & 0.2761 & 0.4290 \\
H-Optimus-1 &  0.6024 & 0.3781 & 0.4964	& 0.6589 & 0.3195 & 0.2421 & 0.2533 & 0.5779 & 0.2774 & 0.4229 \\
GenBio-PathFM & 0.5872 & 0.3913 & 0.4959 & 0.6715 & 0.3284 & 0.1785 & 0.2615 & 0.5787 & \textbf{0.2842} & 0.4197 \\
\rowcolor{shadegray}GenBio-PathFM$^\dagger$ & 0.5867 & \textbf{0.3965} & 0.5066 & 0.6473 & 0.3217 & 0.1961 & 0.2436 & 0.5783 & 0.2831 & 0.4178 \\
\rowcolor{shadegray}\textit{Mascaret} & 0.5920 & 0.3760 & 0.5127 & 0.6442 & \textbf{0.3322} & 0.1964 & 0.2435 & 0.5802 & 0.2727 & 0.4167 \\
H-Optimus-0 & 0.5976 & 0.3848 & 0.4911 & 0.6454 & 0.3086 & 0.2216 & 0.2676 & 0.5590 & 0.2591 & 0.4150 \\
UNI2-h & 0.5898 & 0.3569 & 0.5001 & 0.6606 & 0.3015 & 0.2223 & 0.2640 & 0.5587 & 0.2727 & 0.4141 \\
\rowcolor{shadegray}Virchow2$^\dagger$ & 0.5867 & 0.3900 & 0.5005 & 0.6543 & 0.2940 & 0.2056 & 0.2652 & 0.5656 & 0.2597 & 0.4135 \\
\rowcolor{shadegray}Prov-GigaPath$^\dagger$ & 0.5856 & 0.3730 & 0.5138 & 0.6137 & 0.3214 & 0.1891 & 0.2607 & 0.5663 & 0.2648 & 0.4098 \\
\rowcolor{shadegray}H0-mini$^\dagger$ & 0.5820 & 0.3805 & 0.5070 & 0.6305 & 0.2838 & 0.1874 & 0.2681 & 0.5662 & 0.2626 & 0.4076 \\
\rowcolor{shadegray}AquaViT$^\dagger$ & 0.5928 & 0.3738 & 0.5020 & 0.6232 & 0.2935 & 0.2158 & 0.2378 & 0.5550 & 0.2635 & 0.4064 \\
Virchow & 0.5846 & 0.3378 & 0.5159 & 0.6243 & 0.3079 & 0.1981 & 0.2586 & 0.5664 & 0.2610 & 0.4061 \\
AquaViT & 0.5875 & 0.3814 & 0.4763 & 0.6294 & 0.2984 & 0.2200 & 0.2317 & 0.5510 & 0.2649 & 0.4045 \\
Virchow2 & 0.5971 & 0.3529 & 0.4779 & 0.6402 & 0.2581 & 0.2074 & 0.2719 & 0.5685 & 0.2568 & 0.4034 \\
H0-mini & 0.5862 & 0.3687 & 0.4919 & 0.6012 & 0.2494 & 0.1863 & 0.2670 & 0.5482 & 0.2629 & 0.3958 \\
Midnight-12k & 0.5823 & 0.3370 & 0.4900 & 0.6360 & 0.2908 & 0.1856 & 0.2132 & 0.5577 & 0.2642 & 0.3952 \\
\rowcolor{shadegray}\textit{Phaet} & 0.5630 & 0.3546 & 0.4748 & 0.5985 & 0.2915 & 0.1696 & 0.2696 & 0.5622 & 0.2649 & 0.3943 \\
OpenMidnight & 0.5870 & 0.3590 & 0.4731 & 0.5941 & 0.2728 & 0.1762 & 0.2458 & 0.5534 & 0.2598 & 0.3912 \\
Hibou-L & 0.5701 & 0.2945 & 0.4674 & 0.5817 & 0.3040 & 0.1902 & 0.2657 & 0.5762 & 0.2432 & 0.3881 \\
Prov-GigaPath & 0.5515 & 0.3699 & 0.4746 & 0.5619 & 0.2992 & 0.1961 & 0.2430 & 0.5412 & 0.2500 & 0.3875 \\
UNI & 0.5890 & 0.2943 & 0.4807 & 0.6346 & 0.2614 & 0.1836 & 0.2400 & 0.5464 & 0.2559 & 0.3873 \\
\rowcolor{shadegray}Phikon$^\dagger$ & 0.5571 & 0.3639 & 0.4769 & 0.5615 & 0.2708 & 0.1597 & 0.2430 & 0.5649 & 0.2513 & 0.3832 \\
GPFM & 0.5660 & 0.3423 & 0.4601 & 0.5891 & 0.2480 & 0.1646 & 0.2591 & 0.5472 & 0.2371 & 0.3793 \\
CONCH~1.5 & 0.5440 & 0.3808 & 0.4570 & 0.5517 & 0.2802 & 0.1600 & 0.2176 & 0.5513 & 0.2699 & 0.3792 \\
Phikon-v2 & 0.5408 & 0.3545 & 0.4455 & 0.5554 & 0.2500 & 0.1749 & 0.2659 & 0.5419 & 0.2437 & 0.3747 \\
Kaiko ViT-B/8 & 0.5599 & 0.3611 & 0.4601 & 0.5725 & 0.2683 & 0.1623 & 0.2313 & 0.5183 & 0.2273 & 0.3735 \\
CONCH & 0.5363 & 0.3548 & 0.4468 & 0.5787 & 0.2489 & 0.1602 & 0.2180 & 0.5322 & 0.2507 & 0.3696 \\
Phikon & 0.5327 & 0.3420 & 0.4425 & 0.5355 & 0.2623 & 0.1532 & 0.2423 & 0.5466 & 0.2373 & 0.3660 \\
\end{longtable}
}

%% file: tables/inc/pathorob_full_redacted.tex
{\small\setlength{\tabcolsep}{4pt}
\begin{longtable}{@{}lrrrr@{}}
\caption{PathoROB benchmark: extended leaderboard (robustness index~$\uparrow$). Shaded rows (\textbf{model}$^\dagger$) report fine-tuned results. $^{\mathrm{MP}}$~denotes a base model evaluated with mixed precision; unlabelled rows report results from the official PathoROB leaderboard. Sorted by average robustness index (descending).} \label{tab:pathorob_full} \\
\toprule
\textbf{Model} & \multicolumn{1}{c}{{\strut TCGA}} & \multicolumn{1}{c}{{\strut Cam.}} & \multicolumn{1}{c}{{\strut Tolkach}} & \textbf{Average} \\
\midrule
\endfirsthead
\toprule
\textbf{Model} & \multicolumn{1}{c}{{\strut TCGA}} & \multicolumn{1}{c}{{\strut Cam.}} & \multicolumn{1}{c}{{\strut Tolkach}} & \textbf{Average} \\
\midrule
\endhead
\midrule \multicolumn{5}{r}{\textit{continued\ldots}} \\
\endfoot
\bottomrule
\endlastfoot
Atlas 2 & 0.879	& \textbf{0.940}	& 0.964	& \textbf{0.928} \\
\rowcolor{shadegray}\textit{Mascaret} & \textbf{0.893} & 0.907 & \textbf{0.972} & 0.924 \\
\rowcolor{shadegray}GenBio-PathFM$^\dagger$ & 0.863 & 0.926 & 0.966 & 0.918 \\
\rowcolor{shadegray}Virchow2$^\dagger$ & 0.849 & 0.935 & 0.969 & 0.918 \\
\rowcolor{shadegray}H-Optimus-0$^\dagger$ & 0.856 & 0.933 & 0.961 & 0.917 \\
\rowcolor{shadegray}UNI2-h$^\dagger$ & 0.863 & 0.901 & 0.960 & 0.908 \\
GenBio-PathFM & 0.838 & 0.865 & 0.960 & 0.888 \\
\rowcolor{shadegray}Prov-GigaPath$^\dagger$ & 0.827 & 0.872 & 0.956 & 0.885 \\
\rowcolor{shadegray}AquaViT$^\dagger$ & 0.811 & 0.865 & 0.950 & 0.875 \\
\rowcolor{shadegray}H0-mini$^\dagger$ & 0.811 & 0.842 & 0.950 & 0.868 \\
Virchow2 & 0.822 & 0.806 & 0.955 & 0.861 \\
CONCH~1.5 & 0.832 & 0.774 & 0.951 & 0.852 \\
Atlas & 0.826 & 0.785 & 0.938 & 0.850 \\
Virchow & 0.761 & 0.751 & 0.932 & 0.815 \\
H0-mini & 0.794 & 0.718 & 0.932 & 0.815 \\
CONCH & 0.824 & 0.662 & 0.951 & 0.812 \\
H-Optimus-0 & 0.812 & 0.705 & 0.918 & 0.812 \\
\rowcolor{shadegray}\textit{Phaet} & 0.785 & 0.702 & 0.932 & 0.806 \\
AquaViT$^{\mathrm{MP}}$ & 0.781 & 0.673 & 0.925 & 0.793 \\
Midnight-12k$^{\mathrm{MP}}$ & 0.858 & 0.478 & 0.941 & 0.759 \\
UNI2-h & 0.803 & 0.544 & 0.923 & 0.757 \\
MUSK & 0.727 & 0.467 & 0.928 & 0.707 \\
HIPT & 0.614 & 0.649 & 0.726 & 0.663 \\
Prov-GigaPath & 0.738 & 0.399 & 0.754 & 0.630 \\
\rowcolor{shadegray}Phikon$^\dagger$ & 0.731 & 0.244 & 0.914 & 0.630 \\
Kaiko ViT-B/8 & 0.763 & 0.147 & 0.896 & 0.602 \\
UNI & 0.747 & 0.145 & 0.902 & 0.598 \\
RETCCL & 0.593 & 0.318 & 0.878 & 0.596 \\
CTransPath & 0.652 & 0.106 & 0.872 & 0.543 \\
Kang-DINO & 0.661 & 0.043 & 0.832 & 0.512 \\
RudolfV & 0.587 & 0.184 & 0.695 & 0.489 \\
Phikon & 0.623 & 0.011 & 0.795 & 0.476 \\
Phikon-v2 & 0.619 & 0.019 & 0.768 & 0.469 \\
Ciga & 0.511 & 0.135 & 0.693 & 0.446 \\
\end{longtable}
}